\documentclass{article}
\usepackage{iclr2023_conference,times}

\usepackage[utf8]{inputenc}
\usepackage[T1]{fontenc}
\usepackage[linesnumbered,ruled]{algorithm2e}
\usepackage{times}
\usepackage{latexsym}
\usepackage{comment}
\usepackage{amsmath,amsfonts,amssymb,amsthm,mathtools}
\usepackage{graphicx}
\usepackage{xcolor}
\usepackage{color,soul}
\usepackage{nicefrac}
\usepackage{booktabs}
\usepackage{mathtools}
\usepackage{linguex}
\usepackage{subfigure}
\usepackage{array}
\usepackage{float}
\usepackage{bm}
\usepackage{arydshln}
\usepackage{tikz}
\usepackage{multirow}
\usepackage[pagebackref=true]{hyperref} 
\usepackage{algpseudocode}
\usepackage{cleveref}
\usepackage{subcaption}
\usepackage{annotate-equations}
\usepackage{inconsolata}

\usepackage{xurl}
\usepackage{xfrac,dsfont}
\usepackage{tikz,relsize}
\usepackage{enumitem}
\usetikzlibrary{shapes.geometric, shapes.multipart, arrows, arrows.meta, positioning, calc, chains, backgrounds}
\newcommand{\ifacm}[1]{}

\newcommand{\measure}[1]{\hyperref[sec:measures-of-interpretability]{\emph{#1}}}
\newcommand{\motivation}[1]{\hyperref[sec:motivation-for-interpretability]{\emph{#1}}}
\newcommand{\category}[1]{\hyperref[sec:introduction:abstraction-level]{\emph{#1}}}
\newcommand{\intrinsic}[1]{\hyperref[sec:introduction:intrinsic-v-post-hoc]{\emph{#1}}}
\newcommand{\posthoc}[1]{\hyperref[sec:introduction:intrinsic-v-post-hoc]{\emph{#1}}}
\newcommand{\type}[2]{\hyperref[#1]{\emph{#2}}}
\newcommand{\method}[2]{\hyperref[#1]{\emph{#2}}}
\newcommand{\examplefigurewidth}{\linewidth}
 \renewcommand*\backref[1]{\ifx#1\relax \else (p. #1) \fi}
\usetikzlibrary{mindmap}
\usepackage{amsmath,amsfonts,bm}

\def\eqref#1{equation~\ref{#1}}
\def\1{\bm{1}}

\def\va{{\bm{a}}}
\def\vb{{\bm{b}}}

\def\vh{{\bm{h}}}

\def\vn{{\bm{n}}}

\def\vr{{\bm{r}}}

\def\vu{{\bm{u}}}
\def\vv{{\bm{v}}}
\def\vw{{\bm{w}}}
\def\vx{{\bm{x}}}

\def\vz{{\bm{z}}}

\def\mA{{\bm{A}}}

\def\mL{{\bm{L}}}

\def\mO{{\bm{O}}}

\def\mU{{\bm{U}}}
\def\mV{{\bm{V}}}
\def\mW{{\bm{W}}}
\def\mX{{\bm{X}}}

\DeclareMathAlphabet{\mathsfit}{\encodingdefault}{\sfdefault}{m}{sl}
\SetMathAlphabet{\mathsfit}{bold}{\encodingdefault}{\sfdefault}{bx}{n}

\newcommand{\R}{\mathbb{R}}

\newcommand\numberthis{\addtocounter{equation}{1}\tag{\theequation}}

\definecolor{redcolor}{HTML}{990000}

\newcommand\norm[1]{\left\lVert#1\right\rVert}

\definecolor{darkgreen}{HTML}{138808}
\definecolor{green_drawio}{HTML}{82B366}
\definecolor{dark_green_drawio}{HTML}{557543}
\definecolor{dark_red_drawio}{HTML}{990000}
\definecolor{blue_drawio}{HTML}{6C8EBF}
\definecolor{orange_drawio}{HTML}{D79B00}
\definecolor{red_drawio}{HTML}{990000}
\definecolor{grey_drawio}{HTML}{303030}

\iclrfinalcopy % Uncomment for camera-ready version, but NOT for submission.
\begin{document}

\title{A Primer on the Inner Workings of\\ Transformer-based Language Models}
\author{
\hspace{-4pt}Javier Ferrando$^{1}$\thanks{Correspondence to: \texttt{ jferrandomonsonis@gmail.com}.}~, Gabriele Sarti$^{2}$, Arianna Bisazza$^{2}$, Marta R. Costa-jussà$^3$ \vspace{2pt}\\
\hspace{-4pt}$^1$Universitat Politècnica de Catalunya, $^2$CLCG, University of Groningen, $^3$FAIR, Meta
}
\maketitle
\vspace{-18pt}

\begin{abstract}
\vspace{-8pt}
The rapid progress of research aimed at interpreting the inner workings of advanced language models has highlighted a need for contextualizing the insights gained from years of work in this area. This primer provides a concise technical introduction to the current techniques used to interpret the inner workings of Transformer-based language models, focusing on the generative decoder-only architecture. We conclude by presenting a comprehensive overview of the known internal mechanisms implemented by these models, uncovering connections across popular approaches and active research directions in this area.
\vspace{-7pt}
\end{abstract}

%\tableofcontents

\begin{comment}
\begin{table}[H]
\centering
\resizebox{\examplefigurewidth}{!}{\input{figures/overview-table}}
\caption[Overview of \posthoc{post-hoc} interpretability methods]{A}
\label{tab:overview}
\end{table}
\end{comment}

\begin{comment}
\begin{table}[H]
\centering
\resizebox{\examplefigurewidth}{!}{\input{figures/toc-diagram}}
\caption[Overview of \posthoc{post-hoc} interpretability methods]{Paper layout.}
\label{tab:overview}
\end{table}
\end{comment}

\begin{figure}[H]
 \begin{centering}\includegraphics[width=0.935\textwidth]{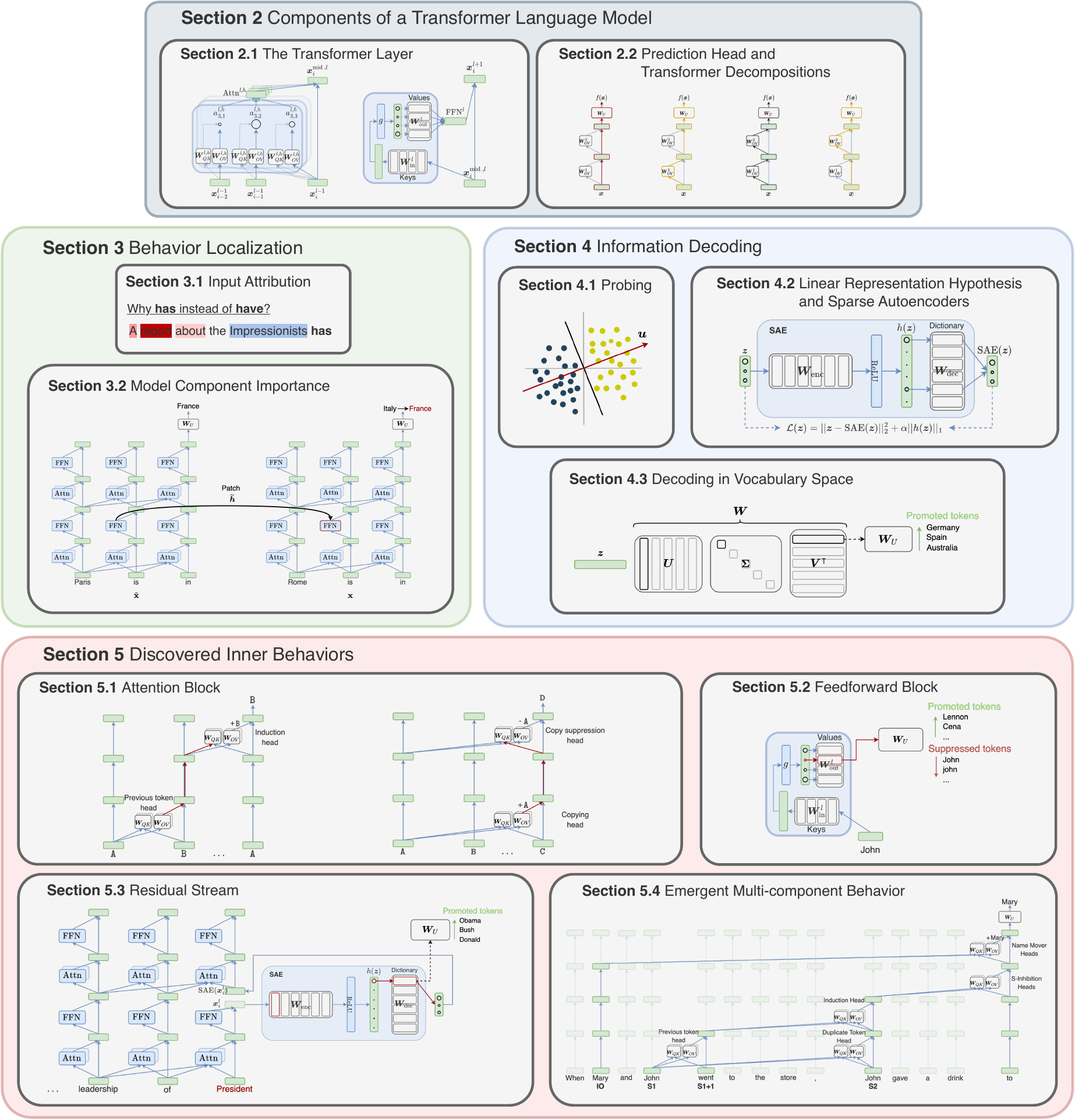}
	\caption{Survey overview. \textbf{\Cref{sec:components_transformer_lm}} introduces the Transformer language 
 model and its components. \textbf{\Cref{sec:behavior_localization}} and \textbf{\Cref{sec:information_decoding}} present interpretability techniques used to analyze models' inner workings. Finally, \textbf{\Cref{sec:what_we_know_transformer}} presents known inner workings of Transformer language models.}
	\label{fig:toc_diagram}
	\end{centering}
\end{figure}

\section{Introduction}
The development of powerful Transformer-based language models~(LMs; \citealp{radford2019language,gpt3,hoffmann2022an,palm}) and their widespread utilization underscores the significance of research devoted to understanding their inner mechanisms. Gaining a deeper understanding of these mechanisms in highly capable AI systems holds important implications in ensuring the safety and fairness of such systems, mitigating their biases and errors in critical settings, and ultimately driving model improvements~\citep{interpretability-safety,costa-jussa-etal-2023-toxicity}. As a result, the natural language processing (NLP) community has witnessed a notable increase in research focused on interpretability in language models, leading to new insights into their internal functioning.

Existing surveys present a wide variety of techniques adopted by Explainable AI analyses~\citep{räuker2023transparent} and their applications in NLP~\citep{posthoc_survey,faithfulness-survey}. While previous NLP interpretability surveys primarily focused on encoder-based models like BERT~\citep{devlin-etal-2019-bert,bertology}, the success of decoder-only Transformers~\citep{radford2018improving} prompted further developments in the analysis of these powerful generative models, with concurrent work surveying trends in interpretability research and their relation to AI safety~\citep{bereska2024mechanistic}. By contrast, this work provides a concise, in-depth technical introduction to relevant techniques used in LM interpretability research, focusing on insights derived from models' inner workings and drawing connections between different areas of interpretability research. Moreover, throughout this work, we employ a unified notation to introduce model components, interpretability methods, and insights from surveyed works, shedding light on the assumptions and motivations behind specific method designs. We categorize LM interpretability approaches surveyed in this work along two dimensions: i)~\textit{localizing} the inputs or model components responsible for a particular prediction (\Cref{sec:behavior_localization}); and ii)~\textit{decoding} information stored in learned representations\footnote{In this work we use \textit{representations} and \textit{activations} interchangeably, and we refer to the fundamental unit of information encoded in model activations as \textit{features}, representing human-interpretable input properties.} to understand its usage across network components (\Cref{sec:information_decoding}). Finally,~\Cref{sec:what_we_know_transformer} provides an exhaustive list of insights into the inner workings of Transformer-based LMs, and~\Cref{sec:lm_interpretability_tools} provides an overview of useful tools to conduct interpretability analyses on these models.

\section{The Components of a Transformer Language Model}\label{sec:components_transformer_lm}

Auto-regressive language models assign probabilities to sequences of tokens. Using the probability chain rule, we can decompose the probability distribution over a sequence ${\mathbf{t} = \langle t_1, t_2 \ldots, t_n \rangle}$ into a product of conditional distributions:
\begin{equation}\label{eq:chain_rule_prob}
P(t_{1},\ldots,t_n) = P(t_{1})\prod_{i=1}^{n-1} P(t_{i+1}|t_{1},\ldots,t_{i}).
\end{equation}
Such distributions can be parametrized using a neural network optimized to maximize the likelihood of a corpus used for training~\citep{bengio_lm}. In recent years, the Transformer architecture by~\citet{vaswani2017transformer} was widely adopted for this purpose thanks to its expressivity and its scalability~\citep{kaplan2020scaling}. While several variants of the original Transformers were proposed,  we focus here on the decoder-only architecture (also known as \textit{GPT-like}) due to its success and popularity.\footnote{Most of the insights presented in this work remain relevant for encoder-only and encoder-decoder models.}
A decoder-only model $f$ has $L$ layers, and operates on a sequence of embeddings ${\mathbf{x} = \langle \vx_1, \vx_2 \ldots, \vx_n \rangle}$ representing the tokens ${\mathbf{t} = \langle t_1, t_2 \ldots, t_n \rangle}$. Each embedding ${\vx \in \R^{d}}$ is a \textit{row vector} corresponding to a row of the embedding matrix $\mW_{E} \in \mathbb{R}^{|\mathcal{V}| \times d}$, where $\mathcal{V}$ is the model vocabulary. Intermediate layer representations, for instance, at position $i$ and layer $l$, are referred to as $\vx^l_i$.\footnote{Note that $\vx^0_i = \vx_i$.} By $\mX \in \mathbb{R}^{n \times d}$ we represent the sequence $\mathbf{x}$ as a matrix with embeddings stacked as rows. Likewise, for intermediate representations, $\mX_{\leq i}^l$ is the layer $l$ representation matrix up to position $i$. \Cref{apx:notation} provides a summary of the notation used in this work.

Following recent literature regarding interpretability in Transformers, we present the architecture adopting the \textit{residual stream} perspective~\citep{elhage2021mathematical}. In this view, each input embedding gets updated via vector additions from the attention~(\Cref{sec:attention_block}) and feed-forward blocks~(\Cref{sec:feedforward_block}), producing \textit{residual stream states} (or intermediate representations). The final layer residual stream state is then projected into the vocabulary space via the unembedding matrix $\mW_U \in \mathbb{R}^{d \times |\mathcal{V}|}$(\Cref{sec:prediction_head}), and normalized via the softmax function to obtain the probability distribution over the vocabulary from which a new token is sampled.

\subsection{The Transformer Layer}\label{sec:transformer_layer}

In this section, we present the Transformer layer components following their computations' flow.

\subsubsection{Layer Normalization}

Layer normalization (LayerNorm) is a common operation used to stabilize the training process of deep neural networks~\citep{ba2016layer}.
Although early Transformer models implemented LayerNorm at the output of each block, modern models consistently normalize preceding each block~\citep{on_layernorm, takase-etal-2023-b2t}. Given a representation $\vz$, the LayerNorm computes $(\nicefrac{(\vz-\mu(\vz))}{\sigma(\vz)}) \odot \mathbf{\gamma}+ \mathbf{\beta}$, where $\mu$ and $\sigma$ calculate the mean and standard deviation, and $\gamma \in \mathbb{R}^{d}$ and $\beta \in \mathbb{R}^{d}$ refer to learned element-wise transformation and bias respectively. Layer normalization can be interpreted geometrically by visualizing the mean subtraction operation as a projection of input representations onto a hyperplane defined by the normal vector ${[1, 1, \ldots, 1] \in \mathbb{R}^{d}}$, and the following scaling to $\sqrt{d}$ norm as a mapping of the resulting representations to a hypersphere~\citep{brody-etal-2023-expressivity,riechers2024geometrydynamicslayernorm}.~\citet{kobayashi-etal-2021-incorporating} notes that LayerNorm can be treated as an affine transformation ${\vz \mL + \beta}$, as long as $\sigma(\vz)$ is considered as a constant~(\Cref{apx:layernorm_linear}). In this view, the matrix $\mL$ computes the centering and scaling operations. Furthermore, the weights of the affine transformation can be folded into the following linear layer (\Cref{appx:folding_ln}), simplifying the analysis.

We note that current LMs such as Llama 2~\citep{touvron2023llama} adopt an alternative layer normalization procedure, RMSNorm~\citep{rms_norm}, where the centering operation is removed, and scaling is performed using the root mean square (RMS) statistic.

\subsubsection{Attention Block}\label{sec:attention_block}

Attention is a key mechanism that allows Transformers to contextualize token representations at each layer. The attention block is composed of multiple \textit{attention heads}.  At a decoding step $i$, each attention head reads from residual streams across previous positions ($\leq i$), decides which positions to attend to, gathers information from those, and finally writes it to the current residual stream. We adopt the rearrangement proposed  by~\citet{kobayashi-etal-2021-incorporating} and~\citet{elhage2021mathematical} to simplify the analysis of residual stream contributions.\footnote{The original implementation considers a concatenation of each attention head output before projecting into the weight matrix $\mW^{l}_{O} \in \mathbb{R}^{H\cdot d_h \times d}$. By splitting $\mW^{l}_O$ into per-head weight matrices $\mW_O^{l,h} \in \mathbb{R}^{d_h \times d}$, matrices $\mW^{l,h}_{V}$ and $\mW^{l,h}_{O}$ can be joined in a single matrix $\mW^{l,h}_{OV}$.} In particular, every attention head computes
\vspace{10pt}
\begin{align*}\label{eq:attn_head_decomposition}
\text{Attn}^{l,h}(\mX^{l-1}_{\le i}) &=  \sum_{j\le i} a^{l,h}_{i,j} \eqnmarkbox[blue_drawio]{v1}{\vx^{l-1}_j \mW^{l,h}_{V}}  \mW^{l,h}_{O}\\
&= \sum_{j\le i} a^{l,h}_{i,j} \vx^{l-1}_j\mW^{l,h}_{OV}.\numberthis
\annotate[yshift=0em]{above, label below}{v1}{Value vector}
\end{align*}
%\vspace{-2pt}
The learnable weight matrices $\mW^{l,h}_V \in \mathbb{R}^{d \times d_h}$ and $\mW^{l,h}_O \in \mathbb{R}^{d_h \times d}$, where $d_h$ represents the dimension of each head, are combined into the OV matrix ${\mW^{l,h}_V \mW^{l,h}_O = \mW^{l,h}_{OV} \in \mathbb{R}^{d \times d}}$, also referred to as \textit{OV (output-value) circuit}. The attention weights for every key ($\leq i$) given the current query ($i$) are obtained as:
\vspace{10pt}
\begin{align*}\label{eq:attention_weights}
\va_{i}^{l,h} &= \text{softmax}\left(\frac{\eqnmarkbox[red_drawio]{qk1}{\vx^{l-1}_i\mW^{l,h}_{Q}} \eqnmarkbox[darkgreen]{qk2}{(\mX^{l-1}_{\le i}\mW^{l,h}_{K})^\intercal}}{\sqrt{d_h}}\right)\\
 &= \text{softmax}\left(\frac{\vx^{l-1}_i\mW^{h}_{QK} {\mX^{l-1}_{\le i}}^{\intercal}}{\sqrt{d_h}}\right),\numberthis
\annotate[yshift=0em]{above, label below}{qk1}{Query vector}
\annotate[yshift=-1em, xshift=4em]{below, label below}{qk2}{Key vectors' matrix}
\end{align*}
%\vspace{-5pt}
with $\mW^{l,h}_Q \in \mathbb{R}^{d \times d_h}$ and $\mW^{l,h}_K \in \mathbb{R}^{d \times d_h}$ combining as the \textit{QK (query-key) circuit} ${\mW^h_{Q} {\mW^{h}_{K}}^{\intercal}= \mW^h_{QK} \in \mathbb{R}^{d \times d}}$. The decomposition introduced in \Cref{eq:attn_head_decomposition,eq:attention_weights} enables a view of QK and OV circuits as units responsible for reading from and writing (in the case of the OV circuit) to the residual stream. The attention block output is the sum of individual attention heads, which is subsequently added back into the residual stream:
\begin{equation}\label{eq:attn_block_decomposition}
\text{Attn}^{l}(\mX^{l-1}_{\le i}) =  \sum\limits_{h=1}^H \text{Attn}^{l,h}(\mX^{l-1}_{\le i}),
\end{equation}
\begin{equation}
    \vx^{\text{mid},l}_i = \vx^{l-1}_i + \text{Attn}^{l}(\mX^{l-1}_{\le i}).
\end{equation}

\begin{figure}[!t]
\begin{centering}\includegraphics[width=0.8\textwidth]{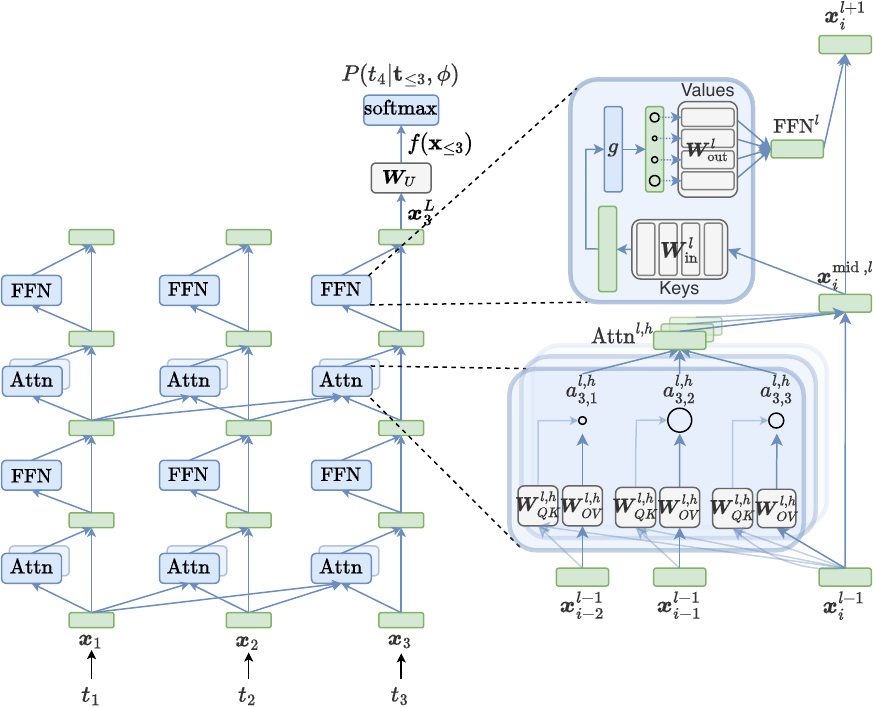}
	\caption{Unrolled Transformer LM with expanded views of the Attention and Feedforward network blocks, including model weights (gray) and residual stream states (green). Based on figures from~\citep{ferrando2024information,voita2023neurons}.}
	\label{fig:attn_ffn_blocks}
	\end{centering}
\end{figure}

\subsubsection{Feedforward Network Block}\label{sec:feedforward_block}
The feedforward network (FFN) in the Transformer block is composed of two learnable weight matrices\footnote{We omit bias terms, following the practice of recent models such as Llama~\citep{touvron2023llama}, PaLM~\citep{palm} and OLMo~\citep{groeneveld2024olmo}, which also exclude biases from attention matrices.}: $\mW^{l}_{\text{in}} \in \mathbb{R}^{d \times d_{\text{ffn}}}$ and $\mW^{l}_{\text{out}} \in \mathbb{R}^{d_{\text{ffn}} \times d}$. $\mW^{l}_{\text{in}}$ reads from the residual stream state $\vx^{\text{mid},l}_i$, and its result is passed through an element-wise non-linear activation function $g$, producing the \textit{neuron activations}. These get transformed by $\mW^l_{\text{out}}$ to produce the output $\text{FFN}(\vx^{\text{mid}}_i)$, which is then added back to the residual stream:
\begin{equation}\label{eq:ffn}
\text{FFN}^{l}(\vx^{\text{mid},l}_i) = g(\vx^{\text{mid},l}_i\mW^l_{\text{in}}) \mW^l_{\text{out}}.
\end{equation}
\begin{equation}
    \vx^{l}_i = \vx^{\text{mid},l}_i + \text{FFN}^{l}(\vx^{\text{mid},l}_i).
\end{equation}

The computation described in \Cref{eq:ffn} was equated to \textit{key-value memory retrieval} \citep{geva-etal-2021-transformer}, with keys ($\vw^l_{\text{in}}$) stored in columns of $\mW^l_{\text{in}}$ acting as pattern detectors over the input sequence (\Cref{fig:attn_ffn_blocks} right) and values $\vw^l_{\text{out}}$, rows of $\mW^l_{\text{out}}$, being upweighted by each neuron activation. We use the term ``neuron'' to refer to each value after an element-wise non-linearity, and use ``unit'' or ``dimension'' for other individual values in any other representation. Provided that the output of the FFN is a linear combination of $\vw^l_{\text{out}}$ values, \Cref{eq:ffn} can be rewritten following the key-value perspective:
\begin{align}\label{eq:key_value_decomposition}
    \text{FFN}^{l}(\vx^{\text{mid},l}_i) &= \sum_{u=1}^{d_{\text{ffn}}} g_u(\vx^{\text{mid},l}_i\vw^{l}_{\text{in}_u}) \vw^{l}_{\text{out}_u}\\
    &= \sum_{u=1}^{d_{\text{ffn}}} n^{l}_u \vw^{l}_{\text{out}_u},\numberthis
\end{align}
with $\vn^{l} \in \mathbb{R}^{d_\text{ffn}}$ being the vector of neuron activations, and $n^{l}_u$ the $u$-th neuron activation value.

The elementwise nonlinearity inside FFNs creates a \textit{privileged basis}~\citep{elhage2022superposition}, which encourages features to align with basis directions. For instance, given a linear network $f(\vx) = \vx \mW_1\mW_2$, the representations extracted from its first layer, $\vx \mW_1$, are rotationally invariant, since we can rotate them by an orthogonal matrix $\mO$, giving $\vx \mW_1 \mO$, and invert the rotation having the output of the network untouched, $f(\vx) = \vx \mW_1\mO \mO^{-1}\mW_2$~\citep{brown2023privileged}. However, having an elementwise nonlinear function on the output of the first layer breaks the rotational invariance of the representations, making the standard basis dimensions (neurons) more likely to be independently meaningful, and therefore better suitable for interpretability analysis.

\subsection{Prediction Head and Transformer Decompositions}\label{sec:prediction_head}
The prediction head of a Transformer consists of an unembedding matrix $\mW_{U} \in \mathbb{R}^{d \times |\mathcal{V}|}$, sometimes accompanied by a bias. The last residual stream state gets transformed by this linear map converting the representation into a next-token distribution of logits, which is turned into a probability distribution via the softmax function.

\paragraph{Prediction as a sum of component outputs.}\label{sec:pred_sum_components} The residual stream view shows that every model component interacts with it through addition~\citep{mickus-etal-2022-dissect}. Thus, the unnormalized scores (logits) are obtained via a linear projection of the summed component outputs. Due to the properties of linear transformations, we can rearrange the traditional forward pass formulation so that each model component contributes directly to the output logits:
\begin{align*}\label{eq:linearization_residual_stream}
f(\mathbf{x})&= \vx^{L}_n\mW_{U}\\
&= \Big(\sum_{l=1}^{L}\sum_{h=1}^{H} \text{Attn}^{l,h}(\mX^{l-1}_{\le n}) + \sum_{l=1}^{L}\text{FFN}^l(\vx^{\text{mid},l}_n) + \vx_n\Big)\mW_{U}\\
&= \sum_{l=1}^{L}\sum_{h=1}^{H} \eqnmarkbox[blue_drawio]{atnn_up}{\text{Attn}^{l,h}(\mX^{l-1}_{\le n})\mW_{U}} + \sum_{l=1}^{L}\eqnmarkbox[green_drawio]{ffn_up}{\text{FFN}^l(\vx^{\text{mid},l}_n)\mW_{U}} + \vx_n\mW_{U}.\numberthis
\annotate[yshift=-0.8em]{below, left, label below}{atnn_up}{Attention head logits update}
\annotate[yshift=-0.9em]{below, right, label below}{ffn_up}{FFN logits update}
\end{align*}

\vspace{4pt}
This decomposition plays an important role when localizing components responsible for a prediction (\Cref{sec:behavior_localization}) since it allows us to measure the direct contribution of every component to the logits of the predicted token~(\Cref{sec:logit_attribution}).
\begin{figure}[!t]
\begin{centering}\includegraphics[width=0.8\textwidth]{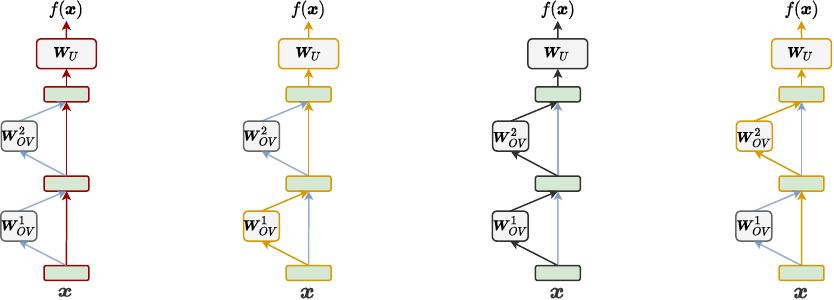}
	\caption{Forward pass decomposition in a simplified Transformer LM. The direct path (red), full OV circuits (yellow) and virtual attention heads (grey) expressed in~\Cref{eq:transformer_paths} are highlighted.\vspace{-5pt}}
	\label{fig:transformer_paths}
	\end{centering}
\end{figure}
\paragraph{Prediction as an ensemble of shallow networks forward passes.}\label{sec:pred_shallow_passes}
Residual networks work as ensembles of shallow networks~\citep{residual_nns_shallow}, where each subnetwork defines a path in the computational graph. Let us consider a two-layer attention-only Transformer, where each attention head is composed just by an OV matrix: ${f(\vx) =\vx^1+\mW_{OV}^2(\vx^1)}$, with ${\vx^1 = \vx + \mW_{OV}^1(\vx)}$. We can decompose the forward pass (\Cref{fig:transformer_paths}) as
%\newpage
\vspace{15pt}
\begin{equation}\label{eq:transformer_paths}
f(\vx) = \eqnmarkbox[red_drawio]{dp}{\vx\mW_U} + \eqnmarkbox[orange_drawio]{fovc1}{\vx\mW_{OV}^1\mW_U} + \eqnmarkbox[grey_drawio]{vcomp}{\vx\mW_{OV}^1 \mW_{OV}^2\mW_U} + \eqnmarkbox[orange_drawio]{fovc2}{\vx\mW_{OV}^2\mW_U}.
\annotate[yshift=0em]{above, left, label below}{dp}{Direct path}
\annotatetwo[yshift=0em]{above, label below}{fovc1}{fovc2}{Full OV circuits}
\annotate[yshift=-0.5em]{below, label above}{vcomp}{Virtual attention heads (V-composition)}
\end{equation}
\vspace{3pt}

The first term in \Cref{eq:transformer_paths}, linking the input embedding to the unembedding matrix, is referred to as the \textit{direct path} (first path in \Cref{fig:transformer_paths}). The paths traversing a single OV matrix are instead named \textit{full OV circuits} (second and fourth path in \Cref{fig:transformer_paths}). Often, full OV circuits are written as $\mW_E \mW_{OV}\mW_U \in \mathbb{R}^{|\mathcal{V}| \times |\mathcal{V}|}$, stacking as rows the logits effect of each input embedding through the circuit. Lastly, the path involving both attention heads is referred to as \textit{virtual attention heads} doing \textit{V-composition}, since the sequential writing and reading of the two heads is seen as OV matrices composing together.~\citet{elhage2021mathematical} propose measuring the amount of composition as: $\nicefrac{\norm{\mW_{OV}^1 \mW_{OV}^2}_F}{\norm{\mW_{OV}^1}_F \norm{\mW_{OV}^2}_F}$. \textit{Q-composition} and \textit{K-composition}, i.e. compositions of $\mW_Q$ and $\mW_K$ with the $\mW_{OV}$ output of previous layers, can also be found in full Transformer models.

\section{Behavior Localization}\label{sec:behavior_localization}
Understanding the inner workings of language models implies localizing which elements in the forward pass (input elements, representations, and model components) are responsible for a specific prediction.\footnote{Commonly referred to as \textit{local explanation} in the interpretability literature~\citep{mythos}}In this section, we present two different types of methods that allow localizing model behavior: \textit{input attribution} (\Cref{sec:input_attribution}) and \textit{model component attribution} (\Cref{sec:components_importance}).

\subsection{Input Attribution}\label{sec:input_attribution}
\textit{Input attribution} methods are commonly used to localize model behavior by estimating the contribution of input elements (in the case of LMs, tokens) in defining model predictions. We refer readers to~\citet{posthoc_survey} for a broader overview of post-hoc input attribution methods with a focus on classification tasks in NLP. Additionally, for a more focused examination of these techniques as applied to Transformer architectures, we recommend the work of~\citet{computers13040092}.

\paragraph{Gradient-based input attribution.} For neural network models like LMs, gradient information is frequently used as a natural metric for attribution purposes~\citep{simonyan_grads,li-etal-2016-visualizing,ding-koehn-2021-evaluating}. \textit{Gradient-based attribution} in this context involves a first-order Taylor expansion of a Transformer at a point $\mathbf{x}$, expressed as ${\nabla f(\mathbf{x}) \cdot\mathbf{x} + \vb}$. The resulting gradient ${\nabla f_w(\mathbf{x}) \in \mathbb{R}^{n \times d} =(\text{grad}\;f_w)(\mathbf{x})}$ captures intuitively the \textit{sensitivity} of the model to each element in the input when predicting token $w$.\footnote{Vocabulary logits or probability scores are commonly used as differentiation targets~\citep{bastings-etal-2022-will}.} While attribution scores are computed for every dimension of input token embeddings, they are generally aggregated at a token level to obtain a more intuitive overview of the influence of individual tokens. This is commonly done by taking the $L^p$ norm of the gradient vector w.r.t the $i$-th input embedding:
\begin{equation}\label{eq:vanilla_gradient}
    A^{\text{Grad}}_{f_{w}(\mathbf{x}) \leftarrow t_i} = \norm{\nabla_{\vx_i}f_w(\mathbf{x})}_p.
\end{equation}
By taking the dot product between the gradient vector and the input embedding ${\nabla_{\vx_i}f_w(\mathbf{x})\cdot \vx_i}$, known as \textit{gradient $\times$ input} method~\citep{denil2015extraction}, this sensitivity can be converted to an importance estimate. However, these approaches are known to exhibit gradient saturation and shattering issues~\citep{gradient_saturation,shattering}. This fact prompted the introduction of methods such as \textit{integrated gradients}~\citep{inte_grad} and SmoothGrad~\citep{smoothgrad} to filter noisy gradient information. For example, integrated gradients approximate the integral of gradients along the straight-line path between a baseline input $\tilde{\mathbf{x}}$ and the input $\mathbf{x}$: ${(\vx_i-\tilde{\vx}_i)\int_{0}^{1}\nabla_{\vx_i}f_w(\tilde{\mathbf{x}} + \alpha(\mathbf{x}-\tilde{\mathbf{x}}))d\alpha}$, and subsequent adaptations were proposed to accommodate the discreteness of textual inputs~\citep{sanyal-ren-2021-discretized,enguehard-2023-sequential}. Finally, approaches based on Layer-wise Relevance Propagation (LRP)~\citep{lrp} have been widely applied to study Transformer-based LMs~\citep{voita-etal-2021-analyzing,chefer-etal-2021-transformer,lrp-transformer,achtibat-etal-2024-attnlrp}. These methods use custom rules for gradient propagation to decompose component contributions at every layer, ensuring their sum remains constant throughout the network.

\begin{figure}[!t]
	\begin{centering}\includegraphics[width=0.8\textwidth]{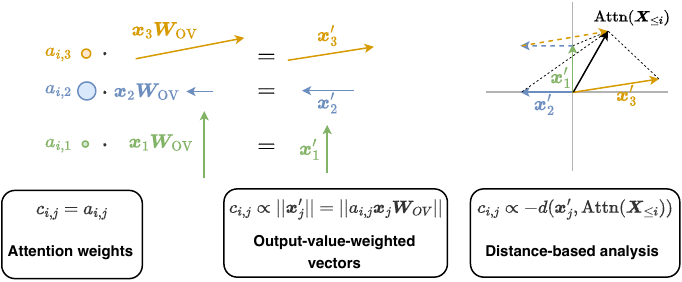}
	\caption{Three approaches to compute inter-token contributions ($c_{i,j}$) towards context mixing in attention heads. Relying only on attention weights overlooks the magnitude of the vectors they operate on. This limitation can be addressed by accounting for the norm of the value-weighted or output-value-weighted vectors ($\vx_j'$). Finally, distance-based analysis estimates the contribution of weighted vectors from their proximity to the attention output.}
	\label{fig:attn_patterns}
	\end{centering}
\end{figure}
\paragraph{Perturbation-based input attribution.} Another popular family of approaches estimates input importance by adding noise or ablating input elements and measuring the resulting impact on model predictions~\citep{li2017understanding}. For instance, the input token at position $i$ can be removed, and the resulting probability difference ${f_w(\mathbf{x}) - f_w(\mathbf{x}_{- \vx_i})}$ can be used as an estimate for its importance. If the logit or probability given to $w$ does not change, we conclude that the $i$-th token has no influence. A multitude of perturbation-based attribution methods exist in the literature, such as those based on \textit{interpretable local surrogate models} such as LIME~\citep{lime}, or those derived from \textit{game theory} like SHAP~\citep{shapley,lundberg-lee-2017-shap}. Notably, new perturbation-based approaches were proposed to leverage linguistic structures~\citep{amara-etal-2024-syntaxshap,zhao2024reagent} and Transformer components~\citep{atman,mohebbi-etal-2023-quantifying} for attribution purposes. These methods relate directly to causal interventions discussed in~\Cref{sec:causal_intervention}. We refer readers to~\citet{explaining-by-removing} for a unified perspective on perturbation-based input attribution.

\paragraph{Context mixing for input attribution.}\label{par:context_mixing} While raw model internals such as attention weights were generally considered to provide unfaithful explanations of model behavior~\citep{jain-wallace-2019-attention,bastings-filippova-2020-elephant,lopardo2024attention}, recent methods have proposed alternatives to attention weights for measuring intermediate token-wise attributions. Some of these alternatives include the use of the norm of value-weighted vectors~\citep{kobayashi-etal-2020-attention} and output-value-weighted vectors~(\citealp{kobayashi-etal-2021-incorporating}), or the use of vectors' distances to estimate contributions~\citep{ferrando-etal-2022-measuring}~(\Cref{fig:attn_patterns} provides a visual description). A common strategy among such approaches involves aggregating intermediate per-layer attributions reflecting \textit{context mixing} patterns~\citep{brunner-etal-2020-identifiability} using techniques such as attention rollout~\citep{abnar-zuidema-2020-quantifying}, resulting in input attribution scores~\citep{ferrando-etal-2022-measuring,modarressi-etal-2022-globenc,mohebbi-etal-2023-quantifying}.\footnote{The attention flow method is seldom used due to its computational inefficiency, despite its theoretical guarantees~\citep{ethayarajh-jurafsky-2021-attention}.} Such context mixing approaches have shown strong faithfulness compared to gradient and perturbation-based methods on classification benchmarks such as ERASER~\citep{deyoung-etal-2020-eraser}. However, rollout aggregation has recently been criticized due to its simplistic assumptions, and recent research has attempted to fully expand the \textit{linear decomposition} of the model output presented in~\Cref{eq:linearization_residual_stream}~\citep{modarressi-etal-2023-decompx,yang-etal-2023-local,oh-schuler-2023-token} as a sum of linear transformations of the input tokens, linearizing the FFN block~\citep{kobayashi2024analyzing}.

\paragraph{Contrastive input attribution.} An important limitation of input attribution methods for interpreting language models is that attributed output tokens belong to a large vocabulary space, often having semantically equivalent tokens competing for probability mass in next-word prediction~\citep{holtzman-etal-2021-surface}. In this context, attribution scores are likely to misrepresent several overlapping factors such as grammatical correctness and semantic appropriateness driving the model prediction. Recent work addresses this issue by proposing a contrastive formulation of such methods, producing counterfactual explanations for why the model predicts token $w$ \textit{instead of} an alternative token $o$~\citep{yin-neubig-2022-interpreting}. As an example, \citet{yin-neubig-2022-interpreting} extend the vanilla gradient method of \Cref{eq:vanilla_gradient} to provide contrastive explanations (ContGrad):
\begin{equation}\label{eq:contrastive_attribution}
    A^{\text{ContGrad}}_{f_{w \neg o}(\mathbf{x}) \leftarrow t_i} = \norm{\nabla_{\vx_i} \left(f_w(\mathbf{x}) - f_o(\mathbf{x}) \right)}_p.
\end{equation}

\paragraph{Limitations of input attribution methods.} While input attribution methods are commonly used to debug failure cases and identify biases in models' predictions~\citep{mccoy-etal-2019-right}, popular approaches were shown to be insensitive to variations in the model and data generating process~\citep{sanity_checks,explanations-lie}, to disagree with each others' predictions~\citep{atanasova-etal-2020-diagnostic,crabbe2023evaluating,krishna2024disagreementproblemexplainablemachine} and to show limited capacity in detecting unseen spurious correlations~\citep{adebayo2020debugging,adebayo2022post}. Importantly, popular methods such as SHAP and Integrated Gradients were found provably unreliable at predicting counterfactual model behavior in realistic settings~\citet{impossibility}. Apart from theoretical limitations, perturbation-based approaches also suffer from out-of-distribution predictions induced by unrealistic noised or ablated inputs, and from high computational cost of targeted ablations for granular input elements.

\paragraph{Training data attribution.}\label{sec:training-data-attribution} Another dimension of input attribution involves the identification of influential training examples driving specific model predictions at inference time~\citep{koh-liang-2017-influence}. These approaches are commonly referred to as \textit{training data attribution} (TDA) or \textit{instance attribution} methods and were applied to identify data artifacts~\citep{han-etal-2020-explaining,pezeshkpour-etal-2022-combining} and sources of biases in language models' predictions~\citep{tda-bias}, with recent approaches proposing to perform TDA via training run simulations~\citep{guu2023simfluence,liu2024training}. While the applicability of established TDA methods was put in question~\citep{akyurek-etal-2022-towards}, especially due to their inefficiency, recent work in this area has produced more efficient methods that can be applied to large generative models at scale~\citep{trak,grosse-etal-2023-studying,kwon2024datainf}. We refer readers to~\citep{hammoudeh2022training} for further details on TDA methods.

\subsection{Model Component Importance}\label{sec:components_importance}
Early studies on the importance of Transformers LMs components highlighted a high degree of sparsity in model capabilities. This means, for example, that removing even a significant fraction of the attention heads in a model may not deteriorate its downstream performances~\citep{sixteen_heads,voita-etal-2019-analyzing}. These results motivated a new line of research studying how various components in an LM contribute to its wide array of capabilities. 

\subsubsection{Logit Attribution}\label{sec:logit_attribution}
\begin{figure}[!t]
	\begin{centering}\includegraphics[width=0.99\textwidth]{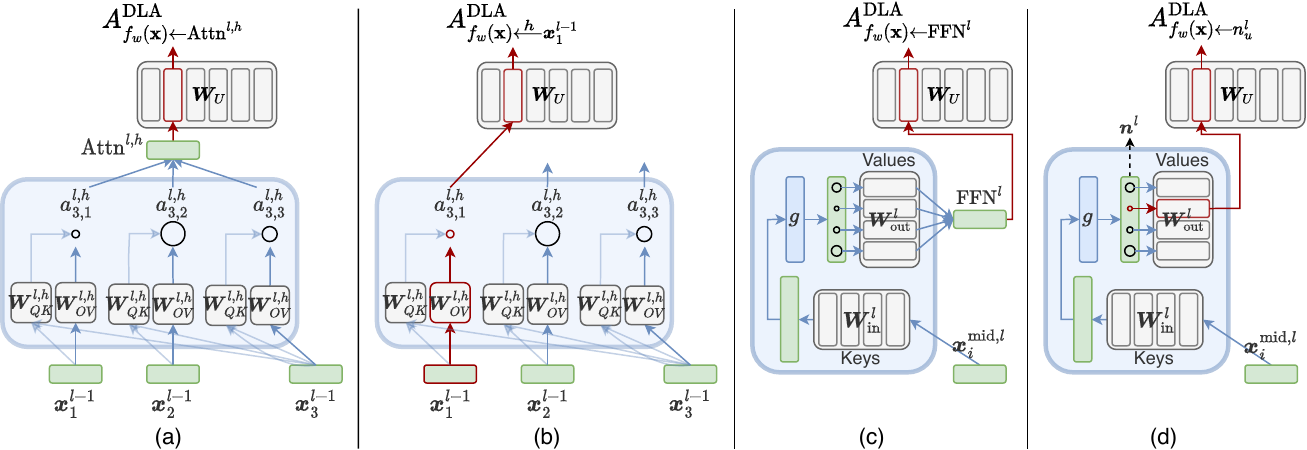}
	\caption{Direct Logit Attributions (DLA) on output token $w$. (a)~DLA of an attention head $\text{Attn}^{l,h}$, (b)~DLA of an intermediate representation $\vx_{1}^{l-1}$ via an attention head, (c)~DLA of an FFN block, and (d)~DLA of a single neuron.}
	\label{fig:dla_attn_ffn}
	\end{centering}
\end{figure}
Let us call $f^{c}(\mathbf{x})$ the output representation of a model component $c$ (attention head or FFN) at a particular layer for the last token position $n$. The decomposition presented in \Cref{eq:linearization_residual_stream} allows us to measure the \textit{direct logit attribution}\footnote{Note that the softmax function is shift-invariant, and therefore the logit scores have no absolute scale.} (DLA,~\Cref{fig:dla_attn_ffn}) of each model component for the output token $w \in \mathcal{V}$:
\begin{equation}\label{eq:dla}
    A^{\text{DLA}}_{f_{w}(\mathbf{x}) \leftarrow c} = f^c(\mathbf{x})\mW_{U[:,w]},%\;\;\text{with}\;\; c \in \{\text{FFN}_{n}^{l}, \text{Attn}_{n}^{l,h}\},
\end{equation}
where $\mW_{U[:,w]}$ is the $w$-th column of $\mW_{U}$, i.e. the unembedding vector of token $w$. In practical terms, the DLA for a component $c$ expresses the contribution of $c$ to the logit of the predicted token, using the linearity of the model's components described in~\Cref{sec:prediction_head}.

\citet{geva-etal-2022-transformer} exploit the fact that the FFN block update is a linear combination of the rows of $\mW_{\text{out}}$ weighted by the neuron activation values (\Cref{eq:key_value_decomposition}). Thus, it is possible to measure the DLA of each neuron as:
\begin{equation}\label{eq:dla_neuron}
    A^{\text{DLA}}_{f_{w}(\mathbf{x}) \leftarrow n^l_u} = n^l_u \vw_{\text{out}_u}^l\mW_{U[:,w]},
\end{equation}
Similarly, \citet{ferrando-etal-2023-explaining} makes use of the decomposition of an attention head as a weighted sum of residual stream transformations~(\Cref{eq:attn_head_decomposition}) and proposes assessing the DLA of each path involving the attention head:
\begin{equation}\label{eq:dla_attention_values}
    A^{\text{DLA}}_{f_{w}(\mathbf{x}) \xleftarrow{\hspace{0.5mm} h \hspace{0.5mm}} \vx^{l-1}_j} = a^{l,h}_{n,j} \vx^{l-1}_j\mW_{OV}^{l,h}\mW_{U[:,w]}.
\end{equation}

The \textit{Logit difference} (LD)~\citep{wang2023interpretability} is the difference in logits between two tokens, ${f_w(\mathbf{x}) - f_o(\mathbf{x})}$. DLA can be extended to measure direct logit difference attribution (DLDA):
\begin{equation}\label{eq:dlda}
    A^{\text{DLDA}}_{f_{w \neg o}(\mathbf{x}) \leftarrow c} = f^c(\mathbf{x})\mW_{U[:,w]} - f^c(\mathbf{x})\mW_{U[:,o]}.
\end{equation}
including its neuron and head-specific variants of \Cref{eq:dla_neuron} and \Cref{eq:dla_attention_values}. Similarly to the contrastive attribution framework described in \Cref{sec:input_attribution}, a positive DLDA value suggests that $c$ promotes token $w$ more than token $o$.

\subsubsection{Causal Interventions}\label{sec:causal_intervention}

\begin{figure}[!t]
	\begin{centering}\includegraphics[width=0.92\textwidth]{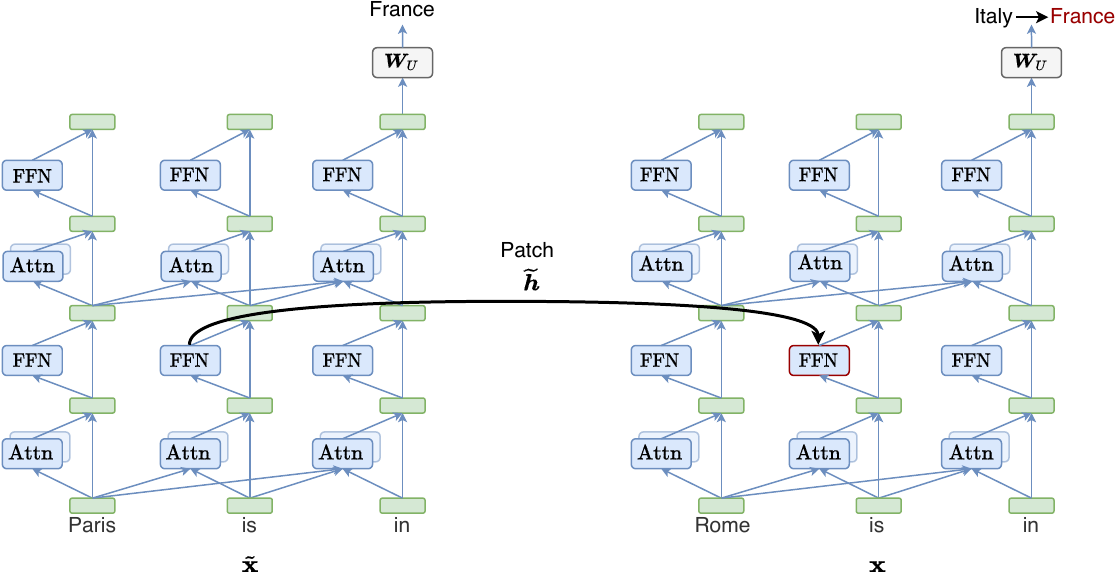}
	\caption{Activation (resample) patching. The FFN output activation from the forward pass with a source input $\tilde{\mathbf{x}}$ (left) is placed in the forward pass with target input $\mathbf{x}$ (right), making the prediction flip from ``Italy'' to ``France''.}
	\label{fig:act_patching}
	\end{centering}
\end{figure}

We can view the computations of a Transformer-based LM as a causal model~\citep{geiger2021causal,mcgrath2023hydra}, and use causality tools~\citep{pearl_2009,causal_mediation_bias} to shed light on the contribution to the prediction of each model component $c \in \mathcal{C}$ across different positions. The causal model can be seen as a directed acyclic graph (DAG), where nodes are model computations and edges are activations. We can intervene in the model by changing some node's value $f^c(\mathbf{x})$ computed by a model component\footnote{Alternatively, we can patch residual stream states $f^l(\mathbf{x})$.} in the forward pass on \textit{target input} $\mathbf{x}$, to those from another value $\tilde{\vh}$, which is referred to as \textit{activation patching}~(\Cref{fig:act_patching}).\footnote{Also referred to in the literature as Causal Mediation Analysis~\citep{causal_mediation_bias}, Causal Tracing~\citep{meng2022locating}, and Interchange Interventions~\citep{geiger-etal-2020-neural,geiger2021causal}.} We can express this intervention using the do-operator~\citep{pearl_2009} as $f(\mathbf{x}|\text{do}(f^c(\mathbf{x})=\tilde{\vh}))$. We then measure how much the prediction changes after patching:
\begin{equation}\label{eq:patching_delta}
A^{\text{Patch}}_{f(\mathbf{x}) \leftarrow c} = \text{diff}(f(\mathbf{x}), f(\mathbf{x}|\text{do}(f^c(\mathbf{x})=\tilde{\vh}))).
\end{equation}
Popular choices for the $\text{diff}(\cdot,\cdot)$ function include KL divergence and logit/probability difference~\citep{zhang2023best}. The patched activation ($\tilde{\vh}$) can be originated from various sources. A common approach is to create a counterfactual dataset with distribution $P_{\text{patch}}$, where some input signals regarding the task are inverted. This approach leads to two distinct types of ablation:
\begin{itemize}
    \item Resample intervention\footnote{Commonly named \textit{ablation} in the literature. We use the more neutral \textit{intervention} here since activations are not actually ablated, but rather replaced.}, where the patched activation is obtained from a single example of $P_{\text{patch}}$, i.e. $\tilde{\vh}=f^c(\tilde{\mathbf{x}}),\tilde{\mathbf{x}} \sim P_{\text{patch}}$~\citep{docstring,hanna2023does,conmy2023automated}.
    \item Mean intervention, where the average of activations of multiple $P_{\text{patch}}$ examples is used for patching, i.e. $\tilde{\vh}=\mathbb{E}_{\tilde{\mathbf{x}} \sim P_{\text{patch}}} [f^c(\tilde{\mathbf{x}})]$~\citep{wang2023interpretability}.
\end{itemize}
Alternatively, other sources of patching activations include:
\begin{itemize}
    \item Zero intervention, where the activation is substituted by a null vector, i.e. $\tilde{\vh}=\mathbf{0}$~\citep{olsson2022context,mohebbi-etal-2023-quantifying}.
    \item Noise intervention, where the new activation is obtained by running the model on a perturbed input, e.g. $\tilde{\vh}=f^c(\mathbf{x} + \epsilon),\epsilon \sim \mathcal{N}(0,\sigma^2)$~\citep{meng2022locating}.
\end{itemize}
An important factor to consider when designing causal interventions experiments is the \textit{ecological validity} of the setup, since zero and noise ablation could lead the model away from the natural activations distribution and ultimately undermine the validity of components' analysis~\citep{chan2022causal,zhang2023best}.

Following the distinction of~\citet{kramár2024atp}, we note that the activation patching methods presented above adopt a \textit{noising} setup, since the patching is performed during the forward pass with the clean/target input, i.e. $f(\mathbf{x}|\text{do}(f^c(\mathbf{x})=\tilde{\vh}))$~\citep{wang2023interpretability,hanna2023does}. Alternatively, the same interventions can be performed in a \textit{denoising} setup, where the patch $\tilde{\vh}$ is taken from the clean/target run and applied over the patched run on source/corrupted input, i.e. $f(\tilde{\mathbf{x}}|\text{do}(f^c(\tilde{\mathbf{x}})=\tilde{\vh}))$~\citep{meng2022locating,lieberum2023does}. We refer readers to~\citet{heimersheim2024use} for a comprehensive overview of metrics, good practices and pitfalls of activation patching.

\textit{Component modeling}~\citep{shah2024decomposing} uses an estimator to measure the effect of interventions on subsets of model components. Specifically,~\citet{shah2024decomposing} demonstrate that a linear function (\textit{component attribution}) can effectively predict the effects of these interventions, providing a clear explanation of how the individual components interact to generate predictions. Other forms of causal interventions use differentiable binary masking on subsets of units or neurons of intermediate representations~\citep{de-cao-etal-2020-decisions,diff_masking,de-cao-etal-2022-sparse}, or entire attention heads outputs~\citep{voita-etal-2019-analyzing,sixteen_heads} which can be cast as a form of zero ablation.

\begin{figure}[!t]
	\begin{centering}\includegraphics[width=0.75\textwidth]{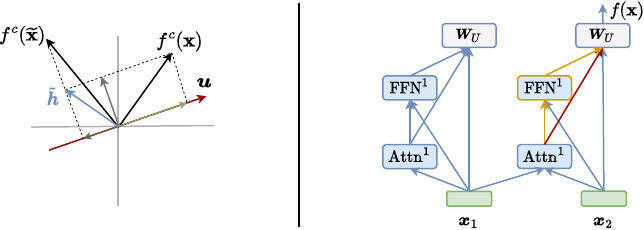}
	\caption{\textbf{Left}: Distributed Interchange Interventions (subspace activation patching) on a 1-dimensional subspace (direction) $\vu$. \textbf{Right}: Single-layer Transformer. Path patching replaces the edges of different paths connecting two nodes (sender and receiver) representing model components. For instance, we can measure the \textcolor{red_drawio}{\textit{direct}} effect of the attention head on the output $f(\mathbf{x})$ or the \textcolor{orange_drawio}{\textit{indirect}} effect of the attention head on the output $f(\mathbf{x})$ \textit{via} the $\text{FFN}$.}
	\label{fig:path_patching_DII_1D}
	\end{centering}
\end{figure}

\paragraph{Subspace Activation Patching.}\label{sec:subspace_patching} It is hypothesized that models encode features as linear subspaces of the representation space~(\Cref{sec:linear_rep_hypothesis}).~\citet{geiger2023finding} proposed distributed interchange interventions (DII), which aim to intervene only on these subspaces.\footnote{Subspace causal interventions were also used as \textit{causal probes} by~\citet{guerner2023geometric}.} It provides a tool that allows for a fine-grained intervention, rather than relying on patching full representations. Formally, assuming a model component $c$ takes values in $\mathbb{R}^{d}$, we seek to find a linear subspace $U \subset \mathbb{R}^{d}$, where by replacing the orthogonal projection of $f^c(\mathbf{x})$ on $U$ with that of $f^c(\mathbf{\widetilde{x}})$ we substitute the feature of interest present in $f^c(\mathbf{x})$ by that in $f^c(\mathbf{\widetilde{x}})$. Following the do-operation notation for the intervention process, $f(\mathbf{x}|\text{do}(f^c(\mathbf{x})=\tilde{\vh}))$, the patched activation is computed as follows:
\begin{equation}
\tilde{\vh} = \underbrace{f^c(\mathbf{x}) - f^c(\mathbf{x})\mU^{\intercal}\mU}_{\text{proj}_{U^{\perp}}f^c(\mathbf{x})} + f^c(\tilde{\mathbf{x}})\mU^{\intercal}\mU
\end{equation}
where $\mU \in \mathbb{R}^{n \times d}$ is an orthonormal matrix whose rows form a basis for $U$. If the feature is encoded as a direction (\Cref{fig:path_patching_DII_1D} left), i.e. in a 1-dimensional subspace, then the patched activation becomes
\vspace{12pt}
\begin{equation}\label{eq:subspace_patching_1D}
\eqnmarkbox[blue_drawio]{patched}{\tilde{\vh}} = \underbrace{f^c(\mathbf{x}) - \eqnmarkbox[green_drawio]{proj_clean}{f^c(\mathbf{x})\vu^{\intercal}\vu}}_{\eqnmarkbox[grey_drawio]{proj_ortho}{\text{proj}_{\vu^{\perp}}f^c(\mathbf{x})}} +\, \eqnmarkbox[dark_green_drawio]{proj_corru}{f^c(\tilde{\mathbf{x}})\vu^{\intercal}\vu}.
\annotate[yshift=-0.2em]{above, left, label below}{patched}{Patched activation}
\annotate[yshift=0em]{above, left, label below}{proj_clean}{Target projection}
\annotate[yshift=0em]{above, right, label below}{proj_corru}{Source projection}
\annotate[yshift=-1em]{below, left, label below}{proj_ortho}{Target projection subtraction}
\end{equation}
\vspace{8pt}

\subsubsection{Circuits Analysis}\label{sec:circuit_analysis}

The Mechanistic Interpretability (MI) subfield focuses on reverse-engineering neural networks into human-understandable algorithms~\citep{olah2022mechinterp}. Recent studies in MI aim to uncover the existence of \textit{circuits}, which are a subset of model components (subgraphs) interacting together to solve a task~\citep{cammarata2020thread}. Activation patching, logit attribution, and attention pattern analysis are common techniques for circuit discovery~\citep{wang2023interpretability,stolfo-etal-2023-mechanistic,stolfo2023understanding,docstring,geva-etal-2023-dissecting,hanna2023does}.

\paragraph{Edge and path patching.}\label{par:path_patching} Activation patching propagates the effect of the intervention throughout the network by recomputing the activations of components after the patched location~(\Cref{fig:act_patching}). The changes in the model output (\Cref{eq:patching_delta}) allow estimating the \textit{total effect} of the model component on the prediction. However, circuit discovery also requires identifying important interactions between components. For this purpose, \textit{edge patching} exploits the fact that every model component input is the sum of the output of previous components in its residual stream~(\Cref{sec:pred_sum_components}), and considers edges directly connecting pairs of model components' nodes~(\Cref{fig:path_patching_DII_1D} right). \textit{Path patching} generalizes the edge patching approach to multiple edges~\citep{wang2023interpretability,goldowskydill2023localizing}, allowing for a more fine-grained analysis. For example, using the forward pass decomposition into shallow networks described in~\Cref{eq:transformer_paths}, we could visualize the single-layer Transformer of~\Cref{fig:path_patching_DII_1D} (right) as being composed as
\vspace{15pt}
\begin{equation}\label{eq:path_patching}
f(\mathbf{x}) = \eqnmarkbox[red_drawio]{adp}{\text{Attn}(\boldsymbol{X}_{\leq n})}\mW_u +\text{FFN}(\eqnmarkbox[orange_drawio]{aip}{\text{Attn}(\mX_{\leq n})} +\vx_n)\mW_u + \vx_n\mW_u,
\end{equation}
\annotate[yshift=0em]{above, left, label below}{adp}{$\text{Attn}$ direct path to logits}
\annotate[yshift=0em]{above, right, label below}{aip}{$\text{Attn}$ indirect path to logits via $\text{FFN}$}
\vspace{-8pt}

where each copy of the \textit{sender} node $\text{Attn}^L(\boldsymbol{X}^{L-1}_{\leq n})$ is relative to a single path. In this example, patching separately each of the sender node copies~\citep{goldowskydill2023localizing} allows us to estimate \textcolor{red_drawio}{\textit{direct}} and \textcolor{orange_drawio}{\textit{indirect}} effects~\citep{pearl_direct_indirect,causal_mediation_bias} of $\text{Attn}^L(\boldsymbol{X}^{L-1}_{\leq n})$ to the output logits $f(\mathbf{x})$. In general, we can apply path patching to any path in the network and measure composition between heads, FFNs, or the effects of these components on the logits.

\paragraph{Limitations of circuit analysis with causal interventions.}\label{par:limitations_circuits} Circuit analysis based on causal intervention methods presents several shortcomings:
\begin{enumerate}
    \item it demands significant efforts for designing the input templates for the task to evaluate, along with the counterfactual dataset, i.e. defining $P_{\text{patch}}$.
    \item isolating important subgraphs after obtaining component importance estimates requires human inspection and domain knowledge.
    \item it has been shown that interventions can produce second-order effects in the behavior of downstream components~(\citealp{makelov2024illusionpatching}, see \citealp{wu2024reply} for discussion), in some settings even eliciting compensatory behavior akin to \textit{self-repair}~\citep{mcgrath2023hydra,rushing2024explorations}. This phenomenon can make it difficult to draw conclusions about the role of each component.
\end{enumerate}

\paragraph{Overcoming the limitations.}\label{par:overcmoing_limitations}~\citet{conmy2023automated} propose an Automatic Circuit Discovery (ACDC) algorithm to automate the process of circuit identification (Limitation 2) by iteratively removing edges from the computational graph. However, this process requires a large amount of forward passes (one per patched element), which becomes impractical when studying large models~\citep{lieberum2023does}. A valid alternative to patching involves gradient-based methods, which have been extended beyond input attribution to compute the importance of intermediate model components ~\citep{influence_explanations,Shrikumar2018ComputationallyEM,dhamdhere2018how}. For instance, given the token prediction $w$, to calculate the attribution of an intermediate layer $l$, denoted as $f^l(\mathbf{x})$, the gradient $\nabla f_w(f^l(\mathbf{x}))$ is computed. ~\citet{inseq} extend the contrastive gradient attribution formulation of \Cref{eq:contrastive_attribution} to locate components contributing to the prediction of the correct continuation over the wrong one using a single forward and backward pass. \citet{attrpatch,syed2023attribution} propose Edge Attribution Patching (EAP), consisting of a linear approximation of the pre- and post-patching prediction difference (\Cref{eq:patching_delta}) to estimate the importance of each edge in the computational graph. The key advantage of this method is that it requires two forward passes and one backward pass to obtain attribution scores of every edge in the graph.~\citet{hanna2024faith} propose combining EAP with Integrated Gradients (EAP-IG) and show improved faithfulness of the extracted circuits, a method also used by~\citet{marks-etal-2024-feature} to identify sparse feature circuits. Further work on Attribution Patching by~\citet{kramár2024atp} finds two settings leading to false negatives in the linear approximation of activation patching, and proposes AtP$^*$, a more robust method preserving a good computational efficiency. Recently,~\citet{ferrando2024information} propose finding relevant subnetworks, which they name \textit{information flow routes}, using a patch-free context mixing approach, requiring only a single forward pass, avoiding the dependence on counterfactual examples and the risk of self-repair interferences during the analysis.

\paragraph{Causal Abstraction.}\label{sec:causal_abstraction}
Another line of research deals with finding interpretable high-level causal abstractions in lower-level neural networks~\citep{geiger2021causal,geiger2022inducing,geiger2023causal}. These methods involve a computationally expensive search and assume high-level variables align with groups of units or neurons. To overcome the limitations,~\citet{geiger2023finding} propose distributed alignment search (DAS), which performs distributed interchange interventions~(DII, \Cref{sec:subspace_patching}) on non-basis-aligned subspaces of the low-level representation space found via gradient descent.\footnote{Alternatively,~\citet{lepori2023uncovering} proposes employing circuit discovery approaches for this purpose.} DAS interventions have been shown to be effective in finding features with causal influence in targeted syntactic evaluation~\citep{arora2024causalgym}, and in isolating the causal effect of individual attributes of entities~\citep{huang2024ravel}. Recently, learned edits on subspaces of intermediate representations during the forward pass have been proposed as an efficient and effective alternative to weight-based Parameter-efficient fine-tuning (PEFT) approaches~\citep{wu2024reft}. A DAS variant named Boundless DAS has been used to search for interpretable causal structure in large language models~\citep{wu-etal-2023-interpretability}. In this context, Causal Proxy Models (CPMs) were proposed as interpretable proxies trained to mimic the predictions of lower-level models and simulate their counterfactual behavior after targeted interventions~\citep{causal_proxy_models}.

\section{Information Decoding}\label{sec:information_decoding}
Fully understanding a model prediction entails localizing the relevant parts of the model, but also comprehending what information is being extracted and processed by each of these components. For example, if the grammatical gender of nouns is assumed to be relevant for the task of coreference resolution in a given language, information decoding methods could look at whether and how a model performing this task encodes noun gender. A natural way to approach decoding the information in the network is in terms of the \textit{features} that are represented in it. 
While there is no universally agreed-upon definition of a feature, it is typically described as a human-interpretable property of the input\footnote{Although we have evidence that models learns human-interpretable features even in instances that exceed human performance~\citep{alphazero_knowledge}, \citet{olah2022mechinterp} argues that the definition of feature should include properties that are not human-interpretable.}, which can be also referred to as a \textit{concept}~\citep{kim_tcav}.

\begin{figure}[!t]
	\begin{centering}\includegraphics[width=0.65\textwidth]{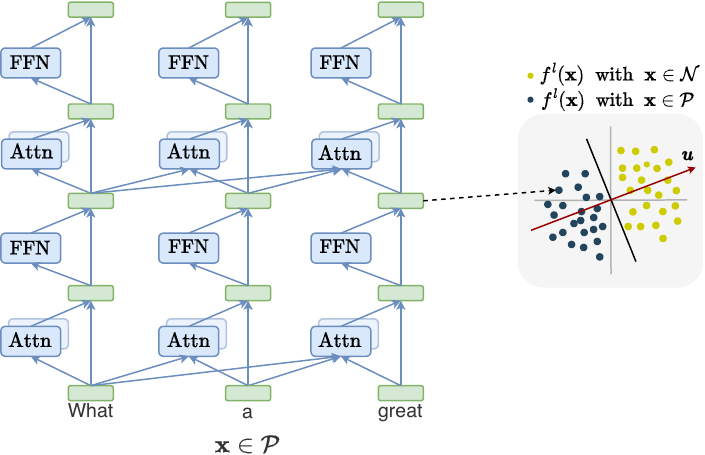}
	\caption{A binary probe trained to predict the input sentiment with positive $\mathcal{P}$ and negative $\mathcal{N}$ sentences. Binary linear classifier probes work within a 1-dimensional subspace (direction $\vu$) in the representation space.\vspace{-5pt}}
	\label{fig:probing}
	\end{centering}
\end{figure}
\subsection{Probing}\label{sec:probing}
Probes, introduced concurrently in NLP by~\citet{kohn-2015-whats,gupta-etal-2015-distributional} and in computer vision by~\citet{alain2016understanding} serve as tools to analyze the internal representations of neural networks. Generally, they take the form of supervised models trained to predict input properties from the representations, aiming to asses how much information about the property is encoded in them. Formally, the probing classifier ${p: f^l(\mathbf{x}) \mapsto z}$ maps intermediate representations to some input features (labels) $z$, which can be, for instance, a part-of-speech tag~\citep{belinkov-etal-2017-neural}, or semantic and syntactic information~\citep{peters-etal-2018-dissecting}. For example, for a binary probe seeking to decode the amount of input sentiment information within an intermediate representation~(\Cref{fig:probing}) we build two sets: $\{f^l(\mathbf{x}): \mathbf{x} \in \mathcal{P}\}$ and $\{f^l(\mathbf{x}): \mathbf{x} \in \mathcal{N}\}$, with the representations obtained when providing positive and negative sentiment sentences respectively. After training the classifier we evaluate the accuracy results on a held-out set.

Although performance on the probing task is interpreted as evidence for the amount of information encoded in the representations, there exists a tension between the ability of the probe to evaluate the information encoded and the probe learning the task itself~\citep{belinkov-2022-probing}. Several works propose using baselines to contextualize the performance of a probe. \citet{hewitt-liang-2019-designing} use \textit{control tasks} by randomizing the probing dataset, while~\citet{pimentel-etal-2020-information} propose measuring the information gain after applying \textit{control functions} on the internal representations.~\citet{voita-titov-2020-information} suggest evaluating the quality of the probe together with the ``amount of effort'' required to achieve the quality. This is done by measuring the minimum description length of the code required to transmit labels $z$ given representations $f^l(\mathbf{x})$. We refer the reader to~\citet{belinkov-glass-2019-analysis,belinkov-2022-probing} for a larger coverage of probing methods.

Probing techniques have been largely applied to analyze Transformers in NLP. Although probes are still being used to study decoder-only models~\citep{chwang2023androids,zou2023representation,burns2023discovering,macdiarmid2024sleeperagentprobes}, a significant portion of the research in this area has focused on BERT~\citep{devlin-etal-2019-bert} and its variants, leading to several BERTology analyses~\citep{bertology}. Probing has provided evidence of the existence of syntactic information within BERT representations~\citep{tenney2018what,lin-etal-2019-open,liu-etal-2019-linguistic}, from which even full parse trees can be recovered with good precision~\citep{hewitt-manning-2019-structural}. Additionally, some studies have analyzed where syntactic information is stored across the residual stream suggesting a hierarchical encoding of language information, with part-of-speech, constituents, and dependencies being represented earlier in the network than semantic roles and coreferents, matching traditional handcrafted NLP pipelines~\citep{tenney-etal-2019-bert}.~\citet{bertology} summarizes results on BERT in detail. Importantly, highly accurate probes indicate a correlation between input representations and labels, but do not provide evidence that the model is using the encoded information for its predictions~\citep{vis_diag_hupkes,belinkov-glass-2019-analysis,amnestic_probing}.

\subsection{Linear Representation Hypothesis and Sparse Autoencoders}\label{sec:linear_rep_hypothesis}

\paragraph{Linear Representation Hypothesis.} The \textit{linear representation hypothesis} states that features are encoded as linear subspaces of the representation space (see~\citet{park2023linear} for a formal discussion).~\citet{mikolov_distributed} were the first to show that Word2Vec word embeddings capture linear syntactic/semantic word relationships. For example, adding the difference between word representations of ``Madrid'' and ``Spain'', $f(``\text{Madrid}")-f(``\text{Spain}")$, to the ``France'' representation, $f(\text{``France''})$, would result in a vector close to $f(\text{``Paris''})$. This presumes that the vector $f(``\text{Madrid}")-f(``\text{Spain}")$ can be considered as the direction of the abstract \textit{capital\_of} feature. Instances of interpretable neurons~\citep{radford2017learning,voita2023neurons,bau_neurons}, i.e. neurons that fire consistently for specific input features (either monosemantic or polysemantic), also exemplify features represented as directions in the neuron space. Recent work suggests the linearity of concepts in representation space is largely driven by the next-word-prediction training objective and inductive biases in gradient descent optimization~\citep{jiang2024origins}.

\paragraph{Erasing Features with Linear Interventions} Feature directions can be found in LMs using linear classifiers (\textit{linear probes}, \Cref{sec:probing}). These models learn a hyperplane that separates representations associated with a particular feature from the rest. The normal vector to that hyperplane, the probe direction $\vu \in \mathbb{R}^{d}$, can be considered the direction representing the underlying feature~(\Cref{fig:probing}). For instance, the sensitivity of model predictions to a feature can be computed as the directional derivative of the model in the direction $\vu$, $\nabla f(f^l(\mathbf{x})) \cdot \vu$, treating the model as a function of the intermediate activation~\citep{kim_tcav}. This linear feature representation was exploited by~\citet{ravfogel-etal-2020-null,linear-erasure,belrose2023leace} to erase concepts, preventing linear classifiers from detecting them in the representation space. Linear concept erasure was shown to mitigate bias~\citep{ravfogel-etal-2020-null} or induce a large increase in perplexity after removing part-of-speech information~\citep{belrose2023leace}. In presence of class labels, linear erasure models can be adapted to ensure the removal of all linear information regarding class identity~\citep{singh2024mimic,belrose2024oracleleace}. Finally,~\citet{amnestic_probing} exploits linear erasure to address the correlational nature of probing classifier, validating the influence of probed properties on model predictions.

\paragraph{Steering Generation with Linear Interventions} As mentioned in \Cref{sec:probing}, a fundamental problem of probing lies in its correlational, rather than causal, nature. Recent work~\citep{nanda-etal-2023-emergent,zou2023representation} shows the effectiveness of linear interventions on language models using directions identified by a probe. For instance, adding negative multiples of the sentiment direction ($\vu$) to the residual stream, i.e. ${\vx^{l'} \leftarrow \vx^l - \alpha \vu}$, is sufficient to generate a text matching the opposite sentiment label~\citep{tigges2023linear}. This simple procedure is named \textit{activation addition}~\citep{turner2023activation}. Other unsupervised methods for computing feature directions include Principal Component Analysis~\citep{tigges2023linear}, K-Means~\citep{zou2023representation}, or difference-in-means~\citep{marks2023geometry}. For instance, \citet{refusal_direction_2024} use the difference-in-means vector between residual streams on harmful and harmless instructions to find a ``refusal direction''~\citep{zheng2024promptdriven} in LMs with safety fine-tuning~\citep{bai2022training}. Projecting out this direction from every model component output, i.e. ${f^{c'} \leftarrow f^{c} - f^{c}\vu^{\intercal}\vu}$, leads to bypass refusal. Recent studies set distributed alignment search~(\Cref{sec:causal_abstraction}) as the best performing method for causal intervention across mathematical reasoning and linguistic plausibility benchmarks~\citep{tigges2023linear,arora2024causalgym,huang2024ravel}, and leveraged it for efficient inference-time interventions aimed at improving task-specific model performance~\citep{wu2024reft}. Finally, the MiMic framework~\citep{singh2024mimic} was recently proposed to craft optimal steering vectors, exploiting insights from linear erasure methods and class labels from the data distribution. We note that the effectiveness of steering approaches involving linear interventions was recently observed to extend to non-Transformer LMs~\citep{paulo2024does}.

\begin{figure}[!t]
	\begin{centering}\includegraphics[width=0.99\textwidth]{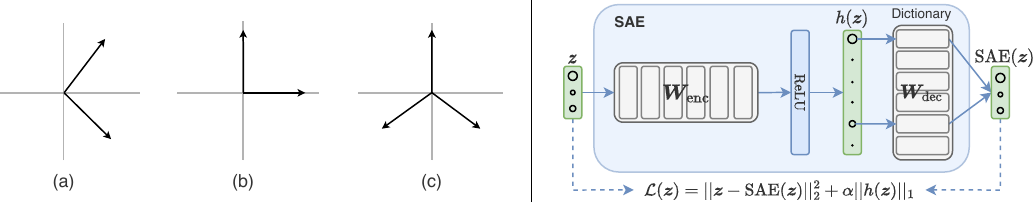}
	\caption{\textbf{Left:} Feature directions in a 2-dimensional space. (a)~features as directions not aligned with the standard basis, we observe polysemanticity. (b)~features aligned with the standard basis, monosemanticity. (c)~more features than dimensions (superposition), hence features can't align with the standard basis and polysemanticity is inevitable. \textbf{Right:} Sparse autoencoder (SAE) trained to reconstruct a model's internal representations $\vz$. Interpretable SAE features are found in rows of $\mW_{\text{dec}}$. Biases are omitted for the sake of clarity.\vspace{-5pt}}
	\label{fig:features_graphs_sae}
	\end{centering}
\end{figure}

\paragraph{Polysemanticity and Superposition.}\label{sec:superposition} A representation produced by a model layer is a vector that lies in a $d$-dimensional space. Neurons are the special subset of representation units right after an element-wise non-linearity (\Cref{sec:feedforward_block}). Although previous work has identified neurons in models corresponding to interpretable features, in most cases they respond to apparently unrelated inputs, i.e. they are \textit{polysemantic}. Two main reasons can explain polysemanticity. Firstly, features can be represented as linear combinations of the standard basis vectors of the neuron space~(\Cref{fig:features_graphs_sae} left (a)), not corresponding to the basis elements themselves. Therefore, each feature is represented across many individual neurons, which is known as \textit{distributed representations}~\citep{smolensky_distributed,olah_distributed}. Secondly, given the extensive capabilities and long-tail knowledge demonstrated by large language models, it has been hypothesized that models could encode more features than they have dimensions, a phenomenon called \textit{superposition}~(\Cref{fig:features_graphs_sae} left (c))~\citep{arora-etal-2018-linear,olah2020zoom}. \citet{elhage2022superposition} showed on toy models trained on synthetic datasets that superposition happens when forcing sparsity on features, i.e. making them less frequent on the training data. Recently, ~\citet{gurnee2023finding} have provided evidence of superposition in the early layers of a Transformer language model, using sparse linear probes.

\paragraph{Sparse Autoencoders (SAEs).}\label{sec:saes} A possible strategy to disentangle features in superposition involves finding an overcomplete feature basis via dictionary learning~\citep{OLSHAUSEN19973311}. Autoencoders with sparsity regularization, also known as \textit{sparse autoencoders} (SAEs), can be used for dictionary learning by optimizing them to reconstruct internal representations $\vz \in \mathbb{R}^{d}$ of a neural network exhibiting superposition while simultaneously promoting feature sparsity. That is, we obtain a reconstruction $\vz = \text{SAE}(\vz) + \epsilon$, where $\epsilon$ is the \textit{SAE error term}.~\citet{sparse_alignment_forum,bricken2023monosemanticity,cunningham2023sparse} propose training SAEs~(see \Cref{fig:features_graphs_sae} right) of the form
\vspace{20pt}
\begin{equation}\label{eq:standard_sae}
    \text{SAE}(\vz) = \eqnmarkbox[green_drawio]{feat_act}{\text{ReLU}\bigl((\vz - \vb_{\text{dec}})\mW_{\text{enc}} + \vb_{\text{enc}})\bigr)}\eqnmarkbox[grey_drawio]{dict}{\mW_{\text{dec}}} + \vb_{\text{dec}}
\annotate[yshift=0em]{above, right, label below}{dict}{Dictionary SAE features}
\end{equation}
\annotate[yshift=0em]{above, left, label below}{feat_act}{SAE feature activations $h(\vz)$}
\hspace{-12pt} on language models' representations with a loss defined as
\vspace{12pt}
\begin{equation}\label{eq:loss_sae}
    \mathcal{L}(\vz) = \eqnmarkbox[red_drawio]{rec_loss}{\norm{\vz - \text{SAE}(\vz)}_2^2} + \eqnmarkbox[green_drawio]{spa_loss}{\alpha\norm{h(\vz)}_1}.
\annotate[yshift=0em]{above, left, label below}{rec_loss}{Reconstruction loss term}
\annotate[yshift=0em]{above, right, label below}{spa_loss}{Sparsity loss term}
\end{equation}
By inducing sparsity on the latent representation of \textit{SAE feature activations} ${h(\vz) = \text{ReLU}(\vz\mW_{\text{enc}} + \vb) \in \mathbb{R}^{m}}$ and setting $m>d$, we can approximate $\vz$ as a sparse linear combination of the rows of the learned ${\mW_{\text{dec}} \in \mathbb{R}^{m \times d}}$ dictionary, from which we can extract interpretable and monosemantic \textit{SAE features}.\footnote{We present in~\Cref{apx:imple_details_saes} some SAEs' implementation details currently debated.} Since the output weights of each SAE feature interact linearly with the residual stream, we can measure their direct effect on the logits~(\Cref{sec:logit_attribution}) and their composition with later layers' components~(\Cref{sec:pred_shallow_passes})~\citep{He2024DictionaryLI}. An initial assessment of reconstruction errors ($\epsilon$) in SAEs trained on LM activations highlighted their systematic nature, driving a shift in next token prediction probabilities much higher than random noise~\citep{sae_errors}.~\citet{marks-etal-2024-feature} also found these errors account for 1–15\% of $\vz$ variance. While this finding might undermine the faithfulness of component analyses relying on SAE features,~\citet{marks-etal-2024-feature} proposes an adaptation of the causal model framework outlined in \Cref{sec:causal_intervention} aiming to incorporate SAE features and errors as nodes of the computational graph. Using edge attribution patching (\Cref{par:overcmoing_limitations}), they recover sparse feature circuits providing more intuitive overviews of features driving model predictions.

\begin{figure}[!t]
	\begin{centering}\includegraphics[width=0.99\textwidth]{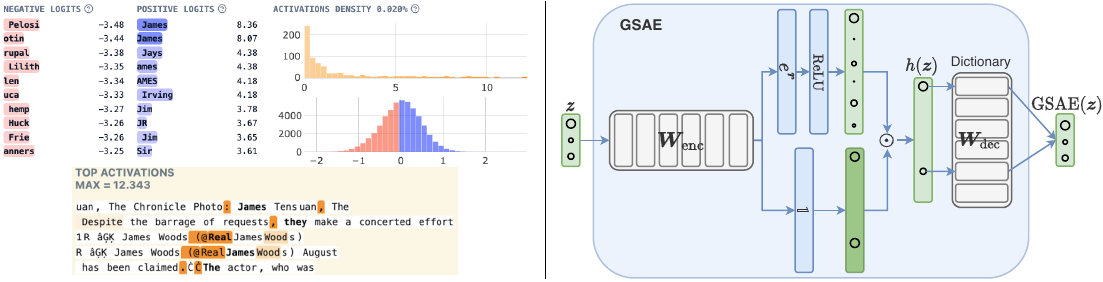}
	\caption{\textbf{Left}: SAE feature visualization on Neuronpedia~\citep{neuronpedia}. It shows the promoted/suppressed tokens, feature density, logits distribution, and maximally activating examples of a \textit{name mover feature} found in GPT-2 Small~\citep{attention_saes_gpt2}. \textbf{Right}: Gated Sparse Autoencoder with encoder weight sharing. Biases are omitted for the sake of clarity.\vspace{-5pt}}
	\label{fig:gated_sae}
	\end{centering}
\end{figure}

\paragraph{SAEs Evaluation.} The goal of SAEs is to learn sparse reconstructions of representations. To assess the quality of a trained SAE in achieving this it is common to compute the \textit{Pareto frontier} of two metrics on an evaluation set~\citep{bricken2023monosemanticity}. These metrics are:
\begin{itemize}
    \item The \textit{L0 norm} of the feature activations vector $h(\vz)$, which measures how many features are ``alive'' given an input. This metric is averaged across the evaluation set, $\mathbb{E}_{\vz\sim \mathcal{D}} \norm{h(\vz)}_0$. 
    \item The \textit{loss recovered}, which reflects the percentage of the original cross-entropy loss of the LM across a dataset when substituting the original representations with the SAE reconstructions.
\end{itemize}
A summary statistic proposed by~\citet{bricken2023monosemanticity} is the \textit{feature density histogram}. \textit{Feature density} is the proportion of tokens in a dataset where a SAE feature has a non-zero value. By looking at the distribution of feature densities we can distinguish if the SAE learnt features that are too dense (activate too often) or too sparse (activate too rarely). Finally, the degree of \textit{interpretability of sparse features} can be estimated based on their \textit{direct logit attribution} and \textit{maximally activating examples}~(see \Cref{fig:gated_sae} left, we introduce these concepts in~\Cref{sec:token_space}). This process can be done manually or automated, using a LLM to produce natural language explanations of SAE features. Although recent studies have compared various SAE architectures~\citep{makelov2024principledevaluationssparseautoencoders,karvonen2024measuringprogressdictionarylearning}, developing robust evaluation frameworks to compare between architectures remains a critical area for future research.

\paragraph{Gated SAEs (GSAEs).} The sparsity penalty used in SAE training promotes smaller feature activations, biasing the reconstruction process towards smaller norms. This phenomenon is known as \textit{shrinkage}~\citep{shrinkage_lasso,feature_suppression}.~\citet{rajamanoharan2024improving} address this issue by proposing Gated Sparse Autoencoders (GSAEs) and a complementary loss function. GSAE is inspired by Gated Linear Units~\citep{gated_cnns,shazeer2020glu}, which employ a gated ReLU encoder to decouple feature magnitude estimation from feature detection~(\Cref{fig:gated_sae} right):
\vspace{9pt}
\begin{equation}\label{eq:gsae}
    \text{GSAE}(\vz) = \underbrace{\eqnmarkbox[dark_green_drawio]{gate}{\mathds{1}[((\vz - \vb_{\text{dec}})\mW_{\text{gate}} + \vb_{\text{gate}})>0]} \odot \eqnmarkbox[green_drawio]{mag}{\text{ReLU}\bigl((\vz - \vb_{\text{dec}})\mW_{\text{mag}} + \vb_{\text{mag}}\bigr)}}_{h(\vz)}\mW_{\text{dec}} + \vb_{\text{dec}},
\annotate[yshift=0em]{above, left, label below}{gate}{GSAE features' gate}
\annotate[yshift=0em]{above, right, label below}{mag}{GSAE feature activations' magnitude}
\end{equation}
% $f_{\text{mag}}(\vz)$, $f_{\text{gate}}(\vz)$
where $\mathds{1}$ is the step function. The features' gate and activation magnitudes are computed by sharing weight matrices, ${\mW_{\text{mag}[i,j]} = \mW_{\text{gate}[i,j]}e^{\vr_{[j]}}}$, being $\vr \in \mathbb{R}^m$ a learned rescaling vector, thus $\mW_{\text{gate}}$ can be considered the encoder matrix $\mW_{\text{enc}}$~(\Cref{fig:gated_sae} right). \citet{rajamanoharan2024improving} show GSAE is a Pareto improvement over the standard SAE architecture on a range of models, scaling GSAEs up to Gemma 7B~\citep{gemmateam2024gemma}.

\paragraph{TopK SAEs.} An alternative approach designed to avoid the L1 penalty is proposed by~\citet{gao2024scaling}, who use k-sparse autoencoders~\citep{makhzani2014ksparse}, a variant that only keeps the $k$ largest features
\begin{equation}\label{eq:topk_saes}
    \text{TopK-SAE}(\vz) = \text{TopK}\bigl((\vz - \vb_{\text{dec}})\mW_{\text{enc}}\bigr)\mW_{\text{dec}} + \vb_{\text{dec}},
\end{equation}
and simplifies the loss to consider only the reconstruction error term (see~\Cref{eq:loss_sae}). TopK SAEs empirically outperform standard ReLU autoencoders~(\Cref{eq:standard_sae}) on the sparsity-reconstruction frontier while increasing monosemanticity. TopK-SAEs constrain each token to use exactly $k$ features. To introduce flexibility in per-sample feature activation,~\citep{bussmann_batch_topk} propose BatchTopK-SAEs, which apply the TopK operation across the flattened batch instead.

\paragraph{JumpReLU SAEs.}~\citet{rajamanoharan2024jumpingaheadimprovingreconstruction} propose a modification to the standard SAE architecture~(\Cref{eq:standard_sae}). Specifically, they substitute the ReLU activation function by the JumpReLU~\citep{erichson2019jumpreluretrofitdefensestrategy}, $\text{JumpReLU}_{\theta}(\vz) = \vz \odot H(\vz - \theta)$, where $H$ is the step function, and $\theta$, a learnable vector acting as a threshold. Intuitively, the output value is zero below the threshold, and the identity afterwards, with a discontinuous jump at the threshold. The threshold $\theta$ enables JumpReLU SAEs to decouple feature activation decisions from the estimation of active feature magnitudes. Additionally, JumpReLUs are trained using the L0 sparsity penalty. Notably,~\citet{lieberum2024gemmascopeopensparse} recently released a set of pre-trained autoencoders on different sites of Gemma 2 models~\citep{gemmateam2024gemma2improvingopen}.

\subsection{Decoding in Vocabulary Space}\label{sec:token_space}

The model engages with the vocabulary in two primary ways: firstly, through a set of input tokens facilitated by the embedding matrix $\mW_E$, and secondly, by interacting with the output space via the unembedding matrix $\mW_U$. Hence, and due to its interpretable nature, a sensible way to approach decoding the information within models' representations is via vocabulary tokens.

\paragraph{Decoding intermediate representations.}\label{sec:res_streams_decoding} The \textit{logit lens}~\citep{nostalgebraist} proposes projecting intermediate residual stream states $\vx^{l}$ by $\mW_U$. The logit lens can also be interpreted as the prediction the model would do if skipping all later layers, and can be used to analyze how the model refines the prediction throughout the forward pass~\citep{jastrzbski2018residual}. This technique has also proven effective in analyzing encoder representations in encoder-decoder models~\citet{langedijk-etal-2023-decoderlens}. However, the logit lens can fail to elicit plausible predictions in some particular models~\citep{belrose2023eliciting}. This phenomenon have inspired researchers to train \textit{translators}, which are functions applied to the intermediate representations prior to the unembedding projection. \citet{din2023jump} suggest using linear mappings, while \citet{belrose2023eliciting} propose affine transformations (\textit{tuned lens}). Translators have also been trained on the outputs of attention heads, resulting in the \textit{attention lens}~\citep{sakarvadia2023attention}. More generally, we can also think of $\mW_U$ as the weights learned by a probe whose classes are the subwords in the vocabulary (\Cref{sec:probing}), and inspect at any point in the network the amount of information encoded about any subword.
%Another variant, named future lens~\citep{pal-etal-2023-future} anticipate the tokens that will appear at future positions.

\paragraph{Patchscopes.} \textit{Patchscopes}~\citep{ghandeharioun2024patchscopes} is a framework that generalizes patching to decode information from intermediate representations.\footnote{A concurrent similar approach is presented by~\citet{chen2024selfie}.} Recall from \Cref{sec:causal_intervention} that patching an activation into a forward pass ${f(\mathbf{x}|\text{do}(f^c(\mathbf{x})=\tilde{\vh}))}$ serves to evaluate the output change with respect to the original clean run ${f(\mathbf{x})}$. Patchscope defines a function acting on the patched representation $m(\tilde{\vh})$, a target model $f^*$ for the patched run, which can differ from the original $f$, a target prompt $\mathbf{x}^*$, and a target model component $c^*$ that can be at a different position and layer. It then evaluates ${f^*(\mathbf{x}^*|\text{do}(f^c(\mathbf{x})^*=m(\tilde{\vh})))}$, either by inspecting the output logits, probabilities, or generating from it a natural language explanation. The election of $f^*$, $m$, $\mathbf{x}^*$, and $c^*$ defines the type of information to extract from $\tilde{\vh}$ independently of the original context, allowing a higher expressivity. For instance, the \textit{future lens}~\citep{pal-etal-2023-future}, used to decode future tokens from intermediate representations can be considered as a patchscope where $f^*=f$, $c^*=c$ and $\mathbf{x}^*$ is a learned prompt.

\begin{figure}[!t]
	\begin{centering}\includegraphics[width=0.99\textwidth]{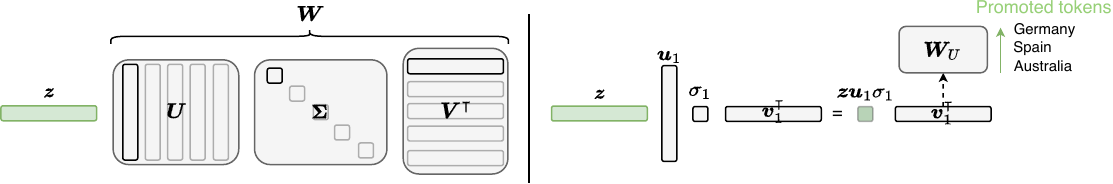}
	\caption{\textbf{Left:} Multiplication of an internal representation $\vz$ by the SVD decomposition of a matrix $\mW$. \textbf{Right:} The top right singular vector $\vv^{\intercal}_1$ represents the direction along which the transformation stretches the most, revealing the tokens the matrix primary interacts with when projecting onto the vocabulary space. The input representation $\vz$ and the associated left singular vector $\vu_1$ act as a query and a key respectively, being $\vv^{\intercal}_1$ the value associated with the key.}
	\label{fig:svd_decomposition}
	\end{centering}
\end{figure}

\paragraph{Decoding model weights.}\label{sec:decoding_model_weights} As seen in previous sections, $\mW^{h}_{OV}$, $\mW_{\text{out}}$, $\mW_{\text{in}}$ and $\mW_{\text{QK}}$ interact linearly with the residual streams. \citet{dar-etal-2023-analyzing} suggest analyzing matrix weights in vocabulary space by projecting them by $\mW_U$, and find that some weight matrices interact with tokens with related semantic meanings. \citet{svd_weights} propose to factorize these matrices via the singular value decomposition (SVD). In the ``thin'' SVD, a matrix is factorized as ${\mW = \mU \mathbf{\Sigma} \mV^{\intercal}}$, with $\mU \in \mathbb{R}^{d \times r}$, $\mathbf{\Sigma} \in \mathbb{R}^{r \times r}$, $\mV^{\intercal} \in \mathbb{R}^{r \times d}$, and $r=\text{rk}(\mW)$, the rank of $\mW$. The largest right singular vectors (rows of $\mV^{\intercal}$)\footnote{Note that we left multiply by $\mW$.} represent the directions along which a linear transformation stretches the most. Then, multiplying $\vz$ by $\mW$~(\Cref{fig:svd_decomposition} left) can be expressed as
\begin{equation}
    \vz \mW = (\vz \mU \mathbf{\Sigma}) \mV^{\intercal}
    = \sum_{i=1}^{r} (\vz \vu_i\sigma_i) \vv_i^{\intercal}.
\end{equation}
where $\vu_i \in \mathbb{R}^{d \times 1}$ can be seen as a key that is compared to the query ($\vz$) via dot product, weighting the right singular vector $\vv_i^{\intercal}$~\citep{intuitions_svd,molina2023traveling}, similar to \Cref{eq:key_value_decomposition}. By projecting the top right singular vectors onto the vocabulary space via the unembedding matrix ($\vv_i^{\intercal} \mW_{U}$) we reveal the tokens the matrix primarily interacts with~(\Cref{fig:svd_decomposition} right). We can instead use the SVD to find a low-rank approximation $\widehat{\mW}(k) = \sum_{i=1}^{k} (\vu_i\sigma_i) \vv_i^{\intercal}$, where $\text{rk}(\widehat{\mW}(k))=k < r$, and study the model predictions by substituting the original matrix by $\widehat{\mW}(k)$~\citep{sharma2024the}. \citet{katz-etal-2024-backward} propose extending the projection of weight matrices~\citep{dar-etal-2023-analyzing} to the backward pass. Specifically, the \textit{backward lens} projects the gradient matrices of the FFNs to study how new information is stored in their weights.
\paragraph{Logit spectroscopy.}\label{logit_spectroscopy} \citet{cancedda2024spectral} proposes an extension of the logit lens, the \textit{logit spectroscopy}, which allows a fine-grained decoding of the information of internal representations via the unembedding matrix ($\mW_u$). Logit spectroscopy considers splitting the right singular matrix of $\mW_u$ into $N$ \textit{bands}: $\{\mV_{u,1}^{\intercal}, \hdots, \mV_{u,N}^{\intercal}\}$, where $\mV_{u,1}^{\intercal}$ and $\mV_{u,N}^{\intercal}$ each contain a set of singular vectors, the former associated with the largest singular values and the latter with the lowest. If we consider the concatenation of matrices associated with different bands, e.g. from the $j$-th to the $k$-th band, we form a matrix $\mV_{u,j:k}^{\intercal}$ whose rows span a linear subspace of the vocabulary space. We can use the operator $\Phi_{u,j:k} = \mV_{u,j:k}\mV_{u,j:k}^{\intercal}$ to evaluate the orthogonal projection $\vz \Phi_{u,j:k}$ of representations $\vz$ onto different subspaces. Alternatively, we can suppress the projection from the representation, i.e. ${\vz' \gets \vz - \vz \Phi_{u,j:k}}$, leaving its orthogonal component with respect to the subspace. Similarly, bands of singular vectors of the embedding matrix can be considered in the analysis.

\paragraph{Maximally-activating inputs.} The features encoded in model neurons or representation units have been largely studied by considering the inputs that maximally activate them~\citep{scenecnn_iclr15,vis_und_cnns}. In image models this can be done either by generating synthesized inputs~\citep{Nguyen_synthesizing}, e.g. via gradient descent~\citep{simonyan_grads}, or by selecting examples from an existing dataset. The latter approach has been used in language models to explain the features that units~\citep{dalvi_grain} and neurons~\citep{neuroscope} respond to. However, \citet{bolukbasi2021interpretability} warn that just relying on maximum activating dataset examples can result in ``interpretability illusions'', as different activation ranges may lead to varying interpretations. Maximally-activating inputs can produce out-of-distribution behaviors, and were recently employed to craft \textit{jailbreak attacks} aimed at eliciting unacceptable model predictions~\citep{chowdhury2024breaking}, for example by crafting maximally-inappropriate inputs for red-teaming purposes~\citep{wichers2024gradientbased}.

\paragraph{Natural language explanations from LMs.} Modern LMs can be prompted to provide plausible-sounding justifications for their own or other LMs' predictions. This can be seen as an edge case of information decoding in which the predictor itself is used as a zero-shot explainer. A notable example is the work by~\citet{openai_neuron_nle} where GPT-4 is prompted to describe shared features in sets of examples producing high activations for specific neurons across GPT-2 XL. Subsequent work by~\citet{huang-etal-2023-rigorously} shows that neurons identified by~\citet{openai_neuron_nle} do not have a causal influence over the concepts highlighted in the generated explanation, underscoring a lack of faithfulness in such approach. Additional investigations in the consistency between input attribution and self-explanations in language models highlighted the tendency of LMs to produce explanations that are very plausible according to human intuition, but unfaithful to model inner workings~\citep{atanasova-etal-2023-faithfulness,parcalabescu-frank-2023-measuring,turpin-etal-2023-language,lanham-etal-2023-measuring,madsen-etal-2024-selfexp,agarwal-etal-2024-faithfulness}.

\section{Discovered Inner Behaviors}\label{sec:what_we_know_transformer}
The techniques presented in \Cref{sec:behavior_localization,sec:information_decoding} have equipped us with essential tools to understand the behavior of language models. In the following sections, we provide an overview of the internal mechanisms that have been discovered within Transformer LMs.

\subsection{Attention Block}\label{sec:insights_attention}
As seen in~\Cref{sec:attention_block}, each attention head consists of a QK (query-key) circuit and an OV (output-value) circuit. The QK circuit computes the attention weights, determining the positions that need to be attended, while the OV circuit moves (and transforms) the information from the attended position into the current residual stream. A substantial body of research has been dedicated to analyzing attention weights patterns formed by QK circuits~\citep{clark-etal-2019-bert,kovaleva-etal-2019-revealing,voita-etal-2019-analyzing}, fueling a debate on whether these weights serve as explanations~\citep{bibal-etal-2022-attention}. However, our understanding of the specific features encoded in the subspaces employed by circuit operations is still limited. Here, we categorize known behavior of attention heads in two groups: those having intelligible attention patterns, and those with meaningful QK and OV circuits.

\subsubsection{Attention heads with interpretable attention weights patterns}

\paragraph{Positional heads.} \citet{clark-etal-2019-bert} showed some BERT heads attend mostly to specific positions relative to the token processed. Specifically, attention heads that attend to the token itself, to the previous token, or to the next position. A similar pattern is also observed in encoders of neural machine translation models~\citep{voita-etal-2019-analyzing,raganato-tiedemann-2018-analysis}. \textbf{Previous token heads} are an essential part of induction heads, and have been shown necessary for circuits in GPT2-Small~\citep{wang2023interpretability}. Their main role has been associated with copying previous token information to the following residual stream, such as concatenating two-tokens names~\citep{nanda2023factfinding}. ~\citet{ferrando2024information} show previous token heads are important across several textual domains.

\paragraph{Subword joiner heads.} First discovered in machine translation encoders~\cite{correia-etal-2019-adaptively}, subword joiner heads have been observed as well in large language models~\citep{ferrando2024information}. These heads attend exclusively to previous tokens that are subwords belonging to the same word as the currently processed token.

\paragraph{Syntactic heads.} Some attention heads attend to tokens having syntactic roles with respect to the processed token significantly more than a random baseline~\citep{clark-etal-2019-bert,htut2019attention}. Particularly, certain heads specialize in given dependency relation types such as \texttt{obj}, \texttt{nsubj}, \texttt{advmod}, and \texttt{amod}.~\citet{angelica2024sudden} show these heads appear suddenly during the training process of masked language models playing a crucial role in the subsequent development of linguistic abilities.

\paragraph{Duplicate token heads.}
Duplicate token heads attend to previous occurrences of the same token in the context of the current token.~\citet{wang2023interpretability} hypothesize that, in the IOI task (\Cref{sec:emergent_behaviors}), these heads copy the position of the previous occurrence to the current position.

\subsubsection{Attention heads with interpretable QK and OV circuits}

\paragraph{Copying heads.} Several attention heads in Transformer LMs have OV matrices that exhibit copying behavior.~\citet{elhage2021mathematical} propose using the number of positive real eigenvalues of the full OV circuit matrix $\mW_E \mW_{OV} \mW_{U}$ as a summary statistic for detecting copying heads. Positive eigenvalues mean that there exists a linear combination of tokens contributing to an increase in the linear combination of logits of the same tokens.

\begin{figure}[!t]
	\begin{centering}\includegraphics[width=0.9\textwidth]{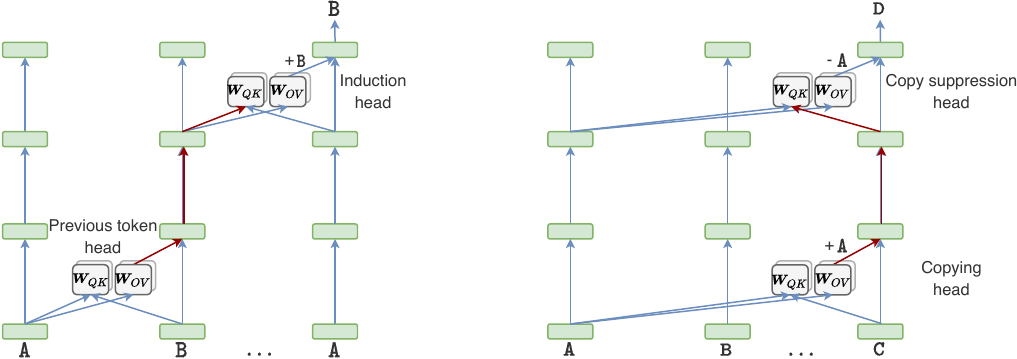}
 \caption{\textbf{Left: Induction mechanism}. An early previous token head writes information of \texttt{A} into \texttt{B}'s residual stream via $\mW_{OV}$. This information (shown in red) gets read by the $\mW_{QK}$ matrix of a downstream induction head (K-composition), which serves it to attend to \texttt{B} and copy its information, increasing the likelihood of \texttt{B} for the next token prediction. \textbf{Right: Copy suppression mechanism}. The copy suppression head detects that a token in the context (\texttt{A}) is being confidently predicted at the current residual stream, for instance, thanks to a previous copying head (shown in red). The copy suppression head attends to it and suppresses its prediction, improving model calibration. Other components are hidden for the sake of clarity.}
	\label{fig:induction_copy_suppression}
	\end{centering}
\end{figure}
\paragraph{Induction heads.}\label{par:induction_heads} An induction mechanism (\Cref{fig:induction_copy_suppression} left) that allows language models to complete patterns was discovered first by~\citet{elhage2021mathematical} and further studied by~\citet{olsson2022context}.\footnote{We follow the mechanistic formulation by~\citet{elhage2021mathematical}. See~\citep{induction-confusion} for a discussion.} This mechanism involves two heads in different layers composing together. Specifically, a previous token head (PTH) and an induction head. The induction mechanism learns to increase the likelihood of token \texttt{B} given the sequence \texttt{A} \texttt{B} ... \texttt{A}, irrespective of what \texttt{A} and \texttt{B} are. To do so, a PTH in an early layer copies information from the first instance of token \texttt{A} to the residual stream of \texttt{B}, specifically by writing in the subspace the QK circuit of the induction head reads from (K-composition). This makes the induction head at the last position to attend to token \texttt{B}, and subsequently, its copying OV circuit increases the logit score of \texttt{B}.~\citet{olsson2022context} demonstrate that the OV and QK circuits of the induction head can perform fuzzy versions of copying and prefix matching, giving rise to generating patterns of the kind $\texttt{A}^*$ $\texttt{B}^*$ ... \texttt{A} $\rightarrow \texttt{B}$, where $\texttt{A}$ and $\texttt{A}^*$, and $\texttt{B}$ and $\texttt{B}^*$ are semantically related (e.g. the same words in different languages). Overall, induction heads have been shown to appear broadly in Transformer LMs~\citep{induction_mosaic}, with those operating at an n-gram level being identified as important drivers of in-context learning~\citep{akyrek2024incontext}. Recent work showed that these heads display both complementary and redundant behaviors, likely shaped by competitive dynamics during optimization (see \Cref{sec:grokking})~\citep{singh2024needs}. Relatedly, redundancy was also observed in the connections between early-layer PTHs and subsequent induction heads. Finally, the emergence rate of induction heads is impacted by the diversity of in-context tokens, with higher diversity in attended and copied tokens delaying the formation of the two respective sub-mechanisms~\citep{singh2024needs}.

\paragraph{Copy suppression heads.}\label{sec:copy_suppression_heads}
Copy suppression heads, discovered in GPT2-Small~\citep{mcdougall2023copy} reduce the logit score of the token they attend to, only if it appears in the context and the current residual stream is confidently predicting it (\Cref{fig:induction_copy_suppression} right).
This mechanism was shown to improve overall model calibration by avoiding naive copying in many contexts (e.g. copying ``love'' in ``\textit{All’s fair in love and \_\_\_}'').
The OV circuit of a copy suppression head can copy-suppress almost all of the tokens in the model’s vocabulary when attended to. This behavior is confirmed by analyzing the ``effective QK circuit'' of GPT2-Small. The key input is the $\text{FFN}^1$ output of every token, and the query input the unembedding of any token, $\mW_U \mW_{QK}\text{FFN}^1(\mW_E)$, and shows the diagonal elements rank higher. Copy suppression is also linked to the self-repair mechanism since ablating an essential component deactivates the suppression behavior, compensating for the ablation.

\paragraph{Successor heads.} Given an input token belonging to an element in an ordinal sequence (e.g. ``one'', ``Monday'', or ``January''), the `effective OV circuit': $\text{FFN}^1(\mW_E) \mW_{OV}\mW_U$ of the successor heads increases the logits of tokens corresponding to the next elements in the sequence (e.g. ``two'', ``Tuesday'', ``February''). Specifically,~\citet{2024successor} show the output of the first FFN block represents a common ‘numerical structure’ on which the successor head acts. \citet{2024successor} find these heads in Pythia~\citep{pythia_models}, GPT2~\citep{radford2019language} and Llama 2~\citep{touvron2023llama} models.

\subsubsection{Other noteworthy attention properties}

\paragraph{Domain specialization.} The attention heads previously described serve specific functions in aiding the model to predict the next token. However, the degree of specialization of components across different domains and tasks remains unclear. \citet{ferrando2024information,chughtai2024summing,lv-etal-2024-interpreting} identify some \textbf{specialized heads} that contribute only within specific input domains, such as non-English contexts, coding sequences, or specific topics. An analysis of the top singular vectors of their OV matrices (\Cref{sec:decoding_model_weights}) reveal these heads mainly promote tokens related to the semantics of the input they participate in.

\begin{figure}[!t]
	\begin{centering}\includegraphics[width=0.8\textwidth]{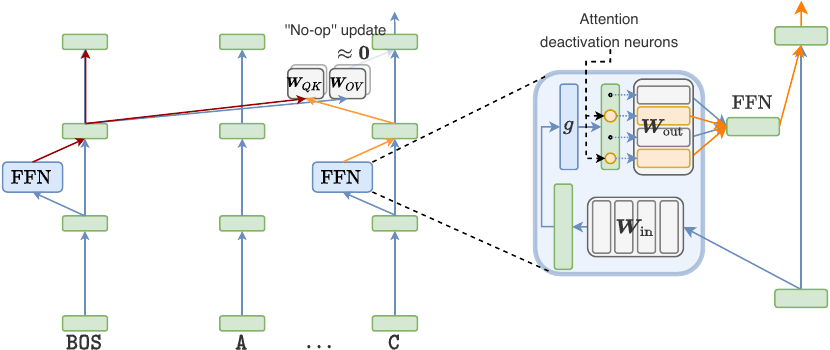}
	\caption{\textbf{Attention sink mechanism}. An attention head at a token $\texttt{C}$ attends to $\texttt{BOS}$ token. Its OV circuit squeezes the $\texttt{BOS}$ residual stream representation resulting in a negligible update, leaving the residual stream of $\texttt{C}$ unchanged. \citet{cancedda2024spectral} suggests that early FFNs in Llama 2 write into a ``dark subspace'' (in red) in the $\texttt{BOS}$ residual stream that allows later heads to exploit this behavior. \citet{gurnee2024universal} find specific neurons in previous layer FFNs of GPT-2 that control the extent to which the query attends to $\texttt{BOS}$ (in orange).}
	\label{fig:attention_sink}
	\end{centering}
\end{figure}
\paragraph{Attention sinks.}\label{sec:attention_sink}
Early investigations into BERT~\citep{kovaleva-etal-2019-revealing} revealed most attention heads exhibit ``vertical'' attention patterns, mainly focusing on special (\texttt{CLS}, \texttt{SEP}) and punctuation tokens. \citet{clark-etal-2019-bert,vig-belinkov-2019-analyzing} hypothesized a head may attend to special tokens when its specialized function is not applicable (no-op hypothesis).~\citet{kobayashi-etal-2020-attention} showed the norm of the value vectors (\Cref{par:context_mixing}) associated with special tokens, periods, and commas tend to be small, canceling out the effect of large attention weights, thereby supporting the no-op hypothesis. Furthermore, it was shown that attention to the end-of-sequence token in MT models is used to ignore the contribution of the source sentence~\citep{ferrando-costa-jussa-2021-attention-weights}, useful when predicting some function words such as the particle ``off'' in \textit{``She turned off the lights.''}. In auto-regressive LMs, these patterns are observed mainly in the beginning of sentence ($\texttt{BOS}$) token~(\Cref{fig:attention_sink}), although other tokens play the same role~\citep{ferrando2024information}. According to~\citet{xiao2023efficient}, allowing attention mass on the $\texttt{BOS}$ token is necessary for streaming generation, and performance degrades when the $\texttt{BOS}$ is omitted. Using the logit spectroscopy~(\Cref{logit_spectroscopy}),
~\citet{cancedda2024spectral} finds that early FFNs in Llama 2 write relevant information (for the attention sink mechanism to occur in later layers) into the residual stream of \texttt{BOS}. These FFNs write into the linear subspace spanned by the right singular vectors with the lowest associated singular values of the unembedding matrix. \citet{cancedda2024spectral} refers to this as a \textit{dark subspace} due to its low interference with next token prediction, and finds a significant correlation between the average attention received by a token and the existence of these dark signals in its residual stream. These dark signals reveal as massive activation values acting as fixed biases~\citep{sun2024massive}, a crucial prerequisite for the attention sink mechanism to take place~\citep{puccetti-etal-2022-outlier,bondarenko2023quantizable}. On the other hand, specific neurons in the FFN of the layer before the attention head have been found to control the amount to which the tokens attend to \texttt{BOS}~\citep{gurnee2024universal}~(\Cref{fig:attention_sink}).

\paragraph{Features in attention heads.} Sparse autoencoders have been trained on the outputs of the attention layers to better understand the features computed by each head. Results presented in~\citet{attention_saes} show that, on a two-layer Transformer, a large number of features (76\%) are non-dead, with the majority of them being interpretable (82\%). Three specific features are studied in detail. The ``\texttt{board} by induction'' feature promotes the token \texttt{board}, and is present on the output of an induction head (\Cref{par:induction_heads}), being part of the \textbf{induction features} family. The ``in questions starting with \texttt{Which}'' feature is instead part of the \textbf{local context features}, promoting the prediction of \texttt{?} when \texttt{Which} appears in the context. Lastly, ``in texts related to pets'' is an example of an \textbf{high-level context feature} that activates for almost the entire context, with its related head attending to pet-related context tokens. Notably, ~\citet{attention_saes} detect the presence of non-induction features in the output of induction-heads, providing evidence of \textbf{attention head polysemanticity}, initially observed by~\citet{docstring}. Further investigations~\citep{attention_saes_gpt2} reveal the same three feature families appear on GPT2-Small, as well as successor features, name mover features, suppression features and duplicate token features associated with heads matching their respective behaviors.~\citet{attention_saes_gpt2_2} conduct a finer-grained analysis of features in GPT-2 Small attention heads, focusing on the top 10 features in each head and concluding that most heads do multiple tasks, with only around 10\% of those being monosemantic. Their findings point out that early layers (0-3) mainly focus on shallow syntactic features, with the following layers encoding increasingly more complex syntactic features. Middle layers (5-6) contain the least interpretable features, while later layers (7-10) encode complex abstract features like time and distance relationships and high-level context concepts. The heads in the last attention block show mostly grammatical adjustments and bigram completions.

\subsection{Feedforward Network Block}\label{sec:insights_ffn}
The dimensions in the FFN activation space (neurons), following the non-linearity, are more likely to be independently meaningful~(\Cref{sec:feedforward_block}), and have therefore been the object of study of recent interpretability works.

\paragraph{Neuron's input behavior.} The behavior of neurons in language models has been extensively studied, with examinations focusing on either their input or output behavior. In the context of input behavior analysis,~\citet{voita2023neurons} show neurons firing exclusively on specific \textbf{position ranges}. Other discoveries include \textbf{skills neurons}, whose activations are correlated with the task of the input prompt~\citep{wang-etal-2022-finding-skill}, \textbf{concept-specific neurons}~\citep{suau2020finding,suau2022selfcond,gurnee2023finding} whose response can be used to predict the presence of a concept in the provided context, such as whether it is Python code, French~\citep{gurnee2023finding}, or German~\citep{quirke2023training} language. Neurons responding to other linguistic and grammatical features have also been found~\citep{bau2018identifying,durrani_ling_neurons}.

\paragraph{Neuron's output behavior.} Regarding the output behavior of neurons, ~\citet{dai-etal-2022-knowledge} use the Integrated Gradients method (\Cref{sec:input_attribution}) to attribute next-word facts predictions to FFNs neurons, finding \textbf{knowledge neurons}. The key-value memory perspective of FFNs (\Cref{sec:feedforward_block}) offers a way to understand neuron's weights. Specifically, using the direct logit attribution method (\Cref{sec:logit_attribution}) we can measure the neuron’s effect on the logits. \citet{geva-etal-2022-transformer} show that some neurons promote the prediction of tokens associated with particular semantic and syntactic concepts. \citet{ferrando-etal-2023-explaining} illustrate that a small set of neurons in later layers is responsible for making \textbf{linguistically acceptable predictions}, such as predicting the correct number of the verb, in agreement with the subject. \citet{gurnee2023language} find neurons that interact with directions in the residual stream that are similar to the \textbf{space and time} feature directions extracted from probes. \citet{tang2024languagespecific} show \textbf{language-specific neurons} are key for multilingual generation, demonstrating one can steer the model output's language by causally intervening on them.~\citet{stolfo2024confidenceregulationneuronslanguage} discover \textbf{token frequency neurons}, which increase or suppress the token’s logit proportionally to its log frequency, shifting the output distribution towards or away from the unigram distribution. Finally, neurons \textbf{suppressing improbable continuations}, e.g. the repetition of the last token in the sequence, have recently been identified~\citep{voita2023neurons,gurnee2024universal}.
 
\paragraph{Polysemantic neurons.} Recent work highlighted the presence of polysemantic neurons within language models. Notably, most early layer neurons specialize in sets of n-grams, functioning as \textbf{n-gram detectors}~\citep{voita2023neurons}, with the majority of neurons firing on a large number of n-grams. \citet{gurnee2023finding} suggest superposition appears in these early layers, and via sparse probing they find sparse combinations of neurons whose added activation values disentangle the detection of specific n-grams, such as the compound word ``social security'' from other bigrams containing only
one of the two terms. Even though polysemanticity and superposition arise in early layers, several \textbf{dead neurons} were observed in OPT models\footnote{OPT models use ReLU activation functions, allowing for zero activation values~\citep{zhang2022opt}.}~\citep{voita2023neurons}. Furthermore, \citet{elhage2022solu} hypothesize models internally perform ``\textbf{de-/re-tokenization}'', where neurons in early layers respond to multi-token words or compound words~\citep{elhage2022solu}, mapping tokens to a more semantically meaningful representation (detokenization). In contrast, in the latest layers, neurons aggregate contextual representations back into single tokens (re-tokenization) to produce the next-token prediction.

\paragraph{Universality of neurons.} Whether different models learn similar features remains an open question~\citep{olah2020zoom}. For instance, various computer vision models were found to learn Gabor filters in early layers~\citep{olah2020an}. In a recent study, \citet{gurnee2024universal} investigated whether neurons respond to features similarly across different models~\citep{bau2018identifying}. Their analysis used the pairwise correlation of neuron activations across GPT2 models trained from different random initializations as a proxy measure, revealing a subset of 1-5\% of neurons activating on the same inputs. As expected, within the cluster of \textbf{universal neurons} there is a higher degree of monosemanticity. This group includes \textbf{alphabet neurons}, which activate in response to tokens representing individual letters and on tokens that start with the letter, supporting the re-tokenization hypothesis. Additionally, there are \textbf{previous token neurons} that fire based on the preceding token, as well as unigram, position, semantic, and syntax neurons. In terms of output behavior, universal neurons include \textbf{attention (de-)activation neurons}, responsible for controlling the amount of attention given to the $\texttt{BOS}$ token by a subsequent attention head, and thus setting it as a no-op~(\Cref{sec:attention_sink}). Lastly, \citet{gurnee2024universal} hypothesize that some neurons act as \textbf{entropy neurons}, modulating the model’s uncertainty over the next token prediction.~\citet{stolfo2024confidenceregulationneuronslanguage} later explained how entropy neurons regulate their confidence by writing to an effective null space of the unembedding matrix and leveraging the layer normalization.

\paragraph{High-level structure of the role of neurons.} It has been suggested that the overall arrangement of neurons in language models mirrors that of neuroscience~\citep{elhage2022solu}. Early layer neurons exhibit similarities to sensory neurons, responding to shallow patterns of the input, mostly focusing on n-grams. Moving into the middle layers, activation tends to occur around more high-level concepts~\citep{bricken2023monosemanticity,gurnee2023finding}. An example of this is the neuron identified in~\citet{elhage2022solu}, which represents numbers only when they refer to the amount of people. Finally, later layers' neurons bear a resemblance to motor neurons in the sense that they produce changes in the distribution of the next-token prediction, either by promoting or suppressing sets of tokens.

\paragraph{Features in Feedforward Networks.}
SAEs are able to identify significantly more interpretable features than the model's neurons themselves~\citep{bricken2023monosemanticity}, as noted both by human and automated analyses in one-layer transformers. The features detected by SAEs trained to reconstruct FFN activations~\citep{bricken2023monosemanticity} appear to split into increasingly more fine-grained distinctions of the feature as more dimensions (dictionary entries) are added, demonstrating that 512 neurons can encode tens of thousands of features. Examples of features found by~\citet{bricken2023monosemanticity} include those firing in the presence of Arabic or Hebrew scripts and promoting tokens in those scripts, and features responding to DNA sequences or base64 strings.

\subsection{Residual Stream}\label{sec:insights_res_stream}
\begin{figure}[!t]
	\begin{centering}\includegraphics[width=0.92\textwidth]{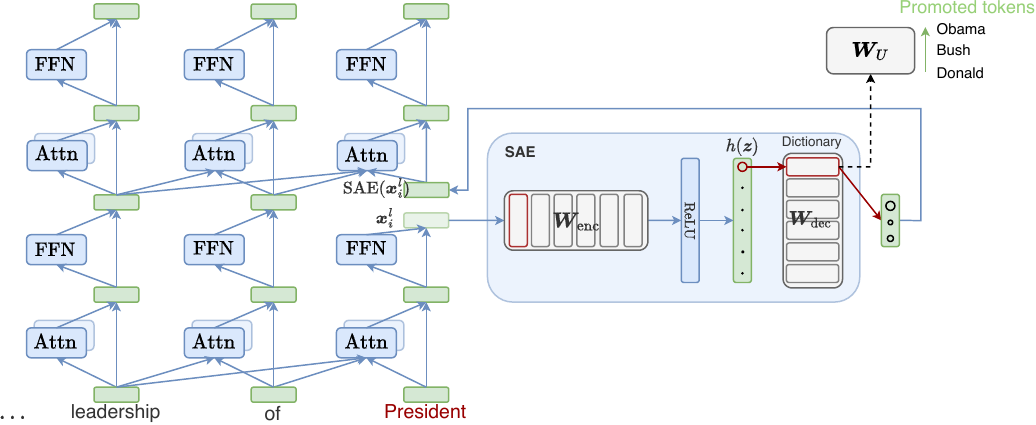}
	\caption{Example of the local context ``President'' feature in a Sparse Autoencoder (SAE) trained to reconstruct the second layer residual stream of GPT2-Small~\citep{bloom2024gpt2residualsaes}.}
	\label{fig:example_sae_feature_president}
	\end{centering}
\end{figure}
% Some info about the residual stream
We can think of the residual stream as the main communication channel in a Transformer. The ``direct path'' (\Cref{sec:prediction_head}) connecting the input embedding with the unembedding matrix, $\vx \mW_U$ does not move information between positions, and mainly models bigram statistics~\citep{elhage2021mathematical}, while the latest biases in the network, localized in the prediction head, are shown to shift predictions according to word frequency, promoting high-frequency tokens~\citep{kobayashi-etal-2023-transformer}. However, alternative paths involve the interaction between components, which write into linear subspaces~\citep{elhage2021mathematical} that can be read by downstream components, or directly by the prediction head, potentially doing more complex computations. \citet{norm_exp_gpt2} observed that \textbf{the norm of the residual stream grows exponentially} along the layers over the forward pass of multiple Transformer LMs~\citep{basic_facts_lms,merrill-etal-2021-effects}. A similar growth rate appears in the norm of the output matrices writing into the residual stream, $\mW_{O}$ and $\mW_{\text{out}}$, unlike input matrices ($\mW_{Q}$, $\mW_{K}$, $\mW_{V}$ and $\mW_{\text{in}}$), which maintain constant norms along the layers. It is hypothesized that \textbf{some components perform \textit{memory management} to remove information stored in the residual stream}. For instance, there are attention heads with OV matrices with negative eigenvalues attending to the current position, and FFN neurons whose input and output weights have large negative cosine similarity~\citep{elhage2021mathematical}, meaning that they write a vector (FFN value) on the opposite direction to the direction they read from (FFN key). Notably, \citet{gurnee2024universal} find that these neurons activate very frequently.~\citet{dao2023adversarial} evaluate a small Transformer LM and provide convincing evidence of multiple attention heads removing the information written by a first layer head.

\textit{Outlier dimensions}~\citep{kovaleva-etal-2021-bert,luo-etal-2021-positional} have been identified within the residual stream. These \textbf{rogue dimensions exhibit large magnitudes relative to others and are associated with the generation of anisotropic representations}~\citep{ethayarajh-2019-contextual,timkey-van-schijndel-2021-bark}. Anisotropy means that the residual stream states of random pairs of tokens tend to point towards the same direction, i.e. the expected cosine similarity is close to one. Furthermore, ablating outlier dimensions has been shown to significantly decrease downstream performance~\citep{kovaleva-etal-2021-bert}, suggesting they encode task-specific knowledge~\citep{rudman-etal-2023-outlier}. The magnitudes of these outliers have been shown to increase with model size~\citep{dettmers2022gptint}, posing challenges for the quantization of large language models. The presence of rogue dimensions has been hypothesized to stem from optimizer choices~\citep{elhage2023privileged}, with higher levels of regularization reducing their magnitudes~\citep{ahmadian2023intriguing}. \citet{puccetti-etal-2022-outlier} identified a high correlation between the magnitude of the outlier dimensions found in token representations and their training frequency. They concluded that \textbf{these dimensions contribute to enabling the model to focus on special tokens, which is known to be associated with ``no-op'' attention updates}~\citep{bondarenko2023quantizable}~(see attention sinks in \Cref{sec:attention_sink}). Recent works have proposed architectural modifications to reduce the presence of these outliers~\citep{bondarenko2023quantizable,hu2024outlier}. In Vision Transformers, high-norm residual stream states have been identified as aggregators of global image information, appearing in patches with highly redundant information, such as those composing the image background~\citep{darcet2024vision}.

% Features found in residual stream
The specific features encoded within the residual stream at various layers remain uncertain, yet sparse autoencoders offer a promising avenue for improving our understanding. Inital SAEs were trained to reconstruct residual stream states of small language models such as GPT2-Small~\citep{cunningham2023sparse,sae_res_stream_logit_lens,bloom2024gpt2residualsaes} showing highly interpretable features~(\Cref{fig:example_sae_feature_president}). Since residual stream states gather information about the sum of previous components' outputs, inspecting SAE's features can illuminate the process by which they are added or transformed during the forward pass. Given the type of features intermediate FFNs and attention heads interact with, we also expect the residual stream at middle layers to encode highly abstract features. \citet{tigges2023linear} provide some preliminary evidence by showing that causally intervening on the residual stream in middle layers is more effective in flipping the sentiment of the output token, suggesting that the latent representation of sentiment is most prominent in the middle layers. \citet{sae_res_stream_logit_lens} study the features learned by a SAE in layer 8 of the 12-layer model GPT2-Small. Based on their output behavior via the logit lens~(\Cref{sec:res_streams_decoding}) the authors first find \textbf{local context features} promoting small sets of tokens. Secondly, they highlight the presence of \textbf{partition features}, which promote and suppress two distinct sets of tokens. For instance, a partition feature might promote tokens starting with capital letters and suppress those starting with lowercase letters. Finally, akin to suppression neurons~\citep{voita2023neurons,gurnee2024universal}, they note the presence of \textbf{suppression features} aimed at reducing the likelihood of specific sets of tokens.~\citet{templeton2024scaling} scale residual stream SAEs to LLMs, finding millions of features in Claude 3 Sonnet~\citep{claude3_sonnet}. They identified \textbf{abstract features}, such as a ``code error feature'' firing on variables incorrectly named and invalid inputs in function calls. Additionally, they found \textbf{safety-related features} on bias or racism, sycophancy features that activate on empathetic text, and others related to deception and power-seeking.~\citet{templeton2024scaling} also showed the model behavior can be steered by clamping the activation values of these features.
%Given a prompt requiring \textbf{multi-step} inference, the most relevant features correspond to intermediate computations
Other notable features in Claude 3 Sonnet include an \textbf{addition function feature} representing the addition operation, similar to function vectors discovered by~\citet{hendel2023incontext,todd2023function}, and the presence of features corresponding to intermediate computations in prompts requiring multi-step inference.
%In the next section, we provide a deeper overview of the interaction between different components and the resulting behavior that emerges.

\subsection{Emergent Multi-component Behaviors}\label{sec:emergent_behaviors}
In previous sections we presented some of the different mechanisms that attention heads and FFNs implement, as well as an overview of the properties of the residual stream. However, in order to explain the remarkable performance of Transformers, we also need to account for the interactions between the different components~\citep{wen-etal-2023-myopic,cammarata2020thread}.

\paragraph{Evidence of multi-component behavior.} The induction mechanism presented in \Cref{par:induction_heads} is a clear example of two components (attention heads) composing together to complete a pattern. Recent evidence suggests that \textbf{multiple attention heads work together to create ``function'' or ``task'' vectors} describing the task when given in-context examples~\citep{hendel2023incontext,todd2023function}. Intervening in the residual stream with those vectors can produce outputs in accordance with the encoded task on novel zero-shot prompts.~\citet{variengien2023look} study in-context retrieval tasks involving answering a request where the answer can be found in the context. The authors identify a high-level mechanism that is universal across subtasks and models. Specifically, middle layers process the request, followed by a retrieval step of the entity from the context done by attention heads at later layers. 

\begin{figure}[!t]	\begin{centering}\includegraphics[width=0.99\textwidth]{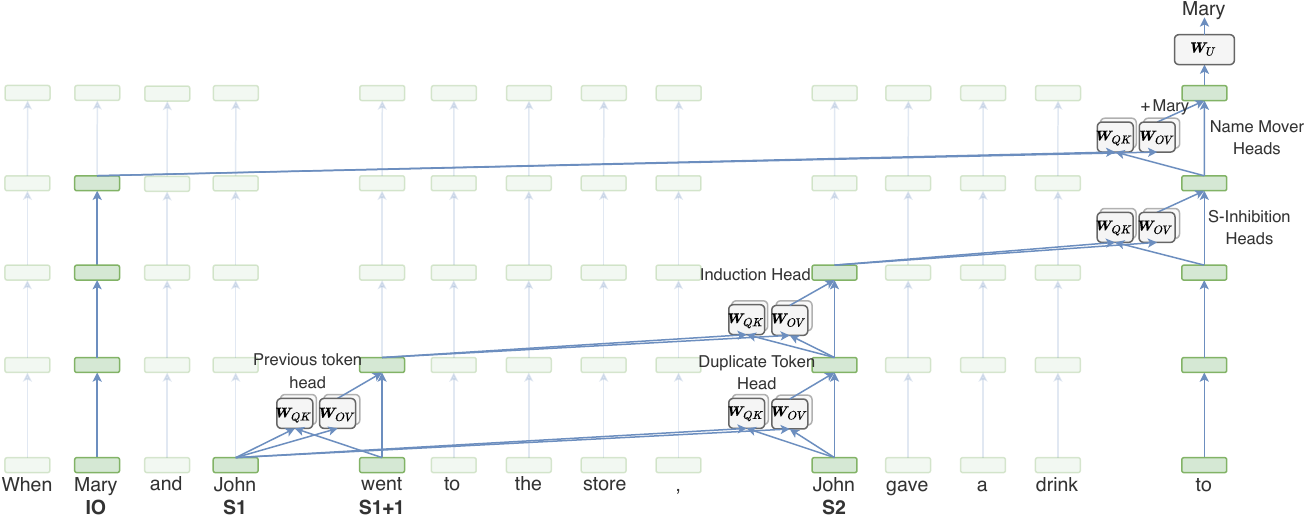}
	\caption{Simplified version of the IOI circuit in GPT2 Small discovered by~\citet{wang2023interpretability}.}
	\label{fig:ioi_circuit}
	\end{centering}
\end{figure}
Additionally, \citet{neo2024interpreting,yu2024locating} reveal that \textbf{individual neurons within downstream FFNs activate according to the output of previous attention heads}, interacting in specific contexts. However, the most compelling evidence of particular behaviors emerging from the interaction between multiple components is found in the circuit analysis literature~(\citet{wang2023interpretability,stolfo2023understanding,docstring,geva-etal-2023-dissecting,hanna2023does}, among others). As an illustration, we present the circuit found in GPT2 Small for the Indirect Object Identification (IOI) task~\citep{wang2023interpretability}, depicted in \Cref{fig:ioi_circuit}. In the IOI task the model is given inputs of the type ``\textit{When Mary and John went to the store, John gave a drink to \_\_\_}''. The initial clause introduces two names (Mary and John), followed by a secondary clause where the two people exchange an item. The correct prediction is the name not appearing in the second clause, referred to as the Indirect Object (Mary). The circuit found in GPT2 Small mainly includes:

\begin{itemize}
    \item Duplicity signaling: duplicate token heads at position S2, and an induction mechanism involving previous token heads at S1+1 signal the duplicity of S (John). This information is read by S-Inhibition heads at the last position, which write in the residual stream a token signal, indicating that S is repeated, and a position signal of the S1 token.
    \item Name copying: name mover heads in later layers copy information from names they attend to in the context to the last residual stream. However, the signals of the previous layers S-Inhibition heads modify the query of name mover heads so that the duplicated name (in S1 and S2) is less attended, favouring the copying of the Indirect Object (IO) and therefore, pushing its prediction.
\end{itemize}
Besides, \citet{wang2023interpretability} discovered Negative mover heads, which are instances of copy suppression heads (\Cref{sec:copy_suppression_heads}) downweighing the probability of the IO. While the IOI is an attention-centric circuit, examples of circuits involving both FFNs and attention heads are also present. For instance, \citet{hanna2023does} reverse-engineered the GPT2-Small circuit for the greater-than task, which involves sentences like \textit{The war lasted from the year 1814 to the year 18\_\_}, where the model must predict a year greater than 1814. The authors demonstrate that downstream FFNs compute a valid year by reading from previous attention heads, which attend to the event's initial date.

\paragraph{Generality of circuits.}~\citet{prakash2024finetuning} show that the \textbf{functionality of the circuit components remains consistent after fine-tuning} and benefits of fine-tuning are largely derived from an improved ability of circuit components to encode important task-relevant information rather than an overall functional rearrangement. Fine-tuned activations are also found to be compatible with the base model despite no explicit tuning constraints, suggesting the process produces minimal changes in the overall representation space. The findings of~\citet{prakash2024finetuning} are additionally supported by~\citet{jain2024what} in controlled settings. While a common critique of mechanistic interpretability work is the limited scope of identified circuits, \citet{merullo2024circuit} show that low-level findings about specific heads and higher-level findings about general algorithms implemented by Transformer models can generalize across tasks, suggesting that large language models could be explained as functions of few task-general sparse components. The results of~\citet{merullo2024circuit} also suggest that circuits are not \textit{exclusive}, i.e. the same model components might be part of several circuits. Other studied dimensions of discovered circuits include their \textit{faithfulness}~\citep{hanna2024faith} and their \textit{completeness}~\citep{wang2023interpretability}.~\citet{ferrando2024similaritycircuitslanguagescase} provide evidence of identical circuits solving subject-verb agreement in both English and Spanish, supporting the generality of discovered circuits across languages.

\paragraph{Grokking as Emergence of Task-specific Circuits.}\label{sec:grokking} Transformer models were observed to converge to different algorithmic solutions for tasks at hand~\citep{zhong-etal-2023-clockpizza}. ~\citet{nanda2023progress} provide convincing evidence on the relation between circuit emergence and \textit{grokking}, i.e. the sudden emergence of near-perfect generalization capabilities for simple symbol manipulation tasks at late stages of model training~\citep{power2022grokking}.~\citet{merril-etal-2023-tale} suggest the grokking phase transition can be seen as the emergence of a sparse circuit with generalization capabilities, replacing a dense subnetwork with low generalization capacity. According to~\citet{Varma2023ExplainingGT}, this happens because dense memorizing circuits are inefficient for compressing large datasets. In contrast, generalizing circuits have a larger fixed cost but better per-example efficiency, hence being preferred in large-scale training.~\citet{huang2024unified} connect the learning dynamic converging to grokking to the \textit{double descent} phenomenon~\citep{doubledescent}. According to this view, the emergence of specialized attention heads might be seen as a mild grokking-related phenomenon~\citep{olsson2022context,bietti-etal-2023-birth}.

\subsubsection{Factuality and hallucinations in model predictions}

\paragraph{Intrinsic views on hallucinatory behavior.}
The generation of factually incorrect or nonsensical outputs is considered a significant limitation in the practical usage of language models~\citep{hallucinations_nlg,minaee2024large}. While some techniques for detecting hallucinated content rely on quantifying the uncertainty of model predictions~\citep{varshney2023stitch}, most alternative approaches engage with model internal representations. Approaches for detecting hallucinations directly from the representations include training probes and analyzing the properties of the representations leading to hallucinations.
~\citet{chwang2023androids} and \citet{azaria-mitchell-2023-internal} find probing classifiers predictive of the model's output truthfulness, achieving the highest accuracy using middle and last layers representations.~\citet{zou2023representation} and \citet{li2023inferencetime} find \textbf{``truthfulness'' directions with causal influence on the model outputs}, i.e. intervening in the internal representations with the found directions enhance the output truthfulness. ~\citet{li2023inferencetime} locate these causal directions in the specific attention head activation.~\citet{chen2024inside} use the eigenvalues of responses' representations covariance matrix to measure the semantic consistency in embedding space across layers, while~\citet{chen2024incontext} observe that logit lens (\Cref{sec:res_streams_decoding}) scores of the predicted attribute (answer) in higher-layers representations of context tokens are informative of the answer correctness.

A related area of research with overlapping goals is that of hallucination detection in machine translation (MT). An MT model is considered to hallucinate if its output contains partially or fully detached content from the source sentence~\citep{guerreiro-etal-2023-looking}. Prediction probabilities of the generated sequence and attention distributions have been used to detect potential errors~\citep{fomicheva-etal-2020-unsupervised} and model hallucinations~\citep{guerreiro-etal-2023-optimal,guerreiro-etal-2023-looking}. Recently, methods measuring the amount of contribution from the source sentence tokens~\citep{ferrando-etal-2022-towards} were found to perform on par with external methods based on semantic similarity across several categories of model hallucinations~\citep{dale-etal-2023-detecting,dale-etal-2023-halomi}. Detection methods show complementary performance across hallucination categories, and simple aggregation strategies for internals-based detectors outperform methods relying on external semantic similarity or quality estimation modules~\citep{himmi-etal-2024-enhanced}.

The underlying mechanisms involved in the prediction of hallucinated content for LLMs remain largely unexplained. Most of the research in this area focuses on studying the ability of language models to recall facts, which we discuss in the next section.

\paragraph{Recall of factual associations.}
\begin{figure}[!t]	\begin{centering}\includegraphics[width=0.68\textwidth]{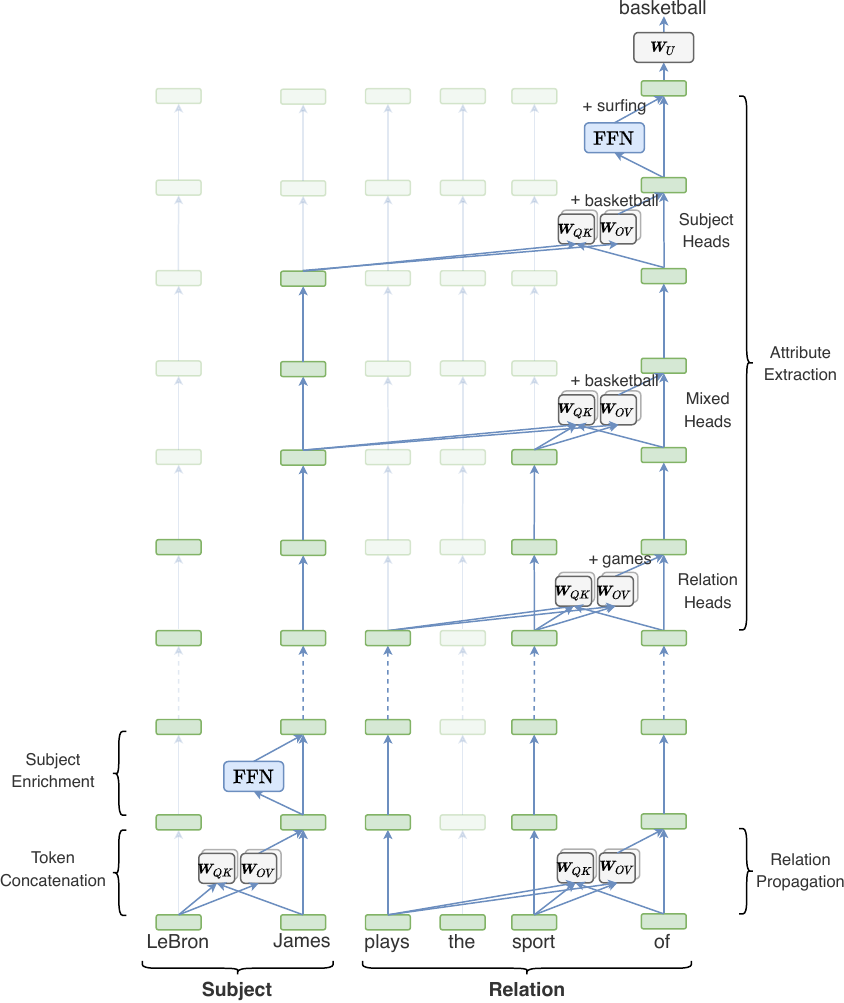}
	\caption{Simplified version of the factual recall circuit.}
	\label{fig:fact_recall_circuit}
	\end{centering}
\end{figure}

Recent research has delved into the internal mechanisms through which language models recall factual information, which is directly related to the hallucination problem in LLMs. A common methodology involves studying tuples $(s,r,a)$, where $s$ is a subject, $r$ a relation, and $a$ an attribute. The model is prompted to predict the attribute given the subject and relation. For instance, given the prompt: \textit{``LeBron James plays the sport of''}, the model is expected to predict \textit{basketball}. \citet{meng2022locating} and \citet{geva-etal-2023-dissecting} make use of causal interventions (\Cref{sec:causal_intervention}) to localize a mechanism responsible for recalling factual knowledge within the language model. \textbf{Early-middle FFNs located in the last subject token add information about the subject} into its residual stream. On the other hand, information from the relation passes into the last token residual stream via early attention heads. Finally, \textbf{later layers attention heads extract the right attribute from the last subject residual stream}. \citet{yuksekgonul2024attention} find that, in similar settings, attention to relevant tokens in the prompt correlates with LLM’s factual correctness. Importantly, the division of responsibilities between lower and upper layers was also observed in attention-less models based on the Mamba architecture~\citep{gu2023mamba,sharma2024locating}. While this might be motivated by implicit context-mixing akin to Transformers' causal self-attention~\citep{ali2024hidden}, it suggests the organization of these mechanisms might be driven by the language modeling optimization process rather than architectural constraints.

Subsequent research has moved from localizing model behavior to studying the computations performed to solve this task. \citet{hernandez2024linearity} show that \textbf{attributes of entities can be linearly decoded from the enriched subject residual stream}, while \citet{chughtai2024summing} investigate how attention heads' OV circuits effectively decode the attributes, proposing an \textbf{additive mechanism}. More precisely, using the direct logit attribution by each token via the attention head (\Cref{eq:dla_attention_values}) they identify subject heads responsible for extracting attributes from the subject independently from the relation (not attending to it), as well as relation heads that promote attributes without being causally dependent on the subject. Additionally, a group of mixed heads generally favor the correct attribute and depend on both the subject and relation. The combination of the different heads' outputs, each proposing different sets of attributes, together with the action of some downstream FFNs resolve the correct prediction (\Cref{fig:fact_recall_circuit}). \citet{nanda2023factfinding} provide a detailed explanation of the subject enrichment phase by studying names of athletes as subjects. They suggest that the \textbf{first layers' attention heads concatenate the athlete's name on the final name token residual stream through addition}, and subsequent FFNs map the obtained athlete's name representation into a linear representation of the athlete's sport that can be easily linearly extracted by the downstream attribute extraction heads.

\citet{merullo2023mechanism} report that for solving relational tasks, such as predicting a country's capital given in-context examples, \textbf{middle layers prepare the argument}, e.g. Poland, \textbf{of a $\texttt{get\_capital}()$ function that is applied downstream via an FFN update}, giving place to ${\texttt{get\_capital}(\text{Poland}) = \text{Warsaw}}$. Further research replicates \citet{merullo2023mechanism}'s analysis on zero-shot settings~\citep{lv-etal-2024-interpreting} and finds specific attention heads ``passing'' the argument from the context (Poland), but also promoting the capital cities (Warsaw). Downstream FFNs ``activate'' relevant attention heads in the previous layer and add a vector guiding the residual stream toward the correct capital direction.

Recent works aim to shed light on how the model engages in factual recall vs. grounding. Following the aforementioned (subject, relation, attribute) structure of facts, an answer is considered to be grounded if the attribute is consistent with the information in the context of the prompt. Given prompts of the type \textit{``The capital of Poland is London. Q: What is the capital of Poland? A:\_\_\_''}, ~\citet{yu-etal-2023-characterizing} find \textbf{in-context heads} and \textbf{memory heads} by using the difference logit attribution (\Cref{sec:logit_attribution}, \Cref{eq:dlda}) of attention heads. These heads favor, respectively, the in-context answer \textit{London} and the memorized answer \textit{Warsaw}, showing a ``competition'' between mechanisms~\citep{ortu-etal-2024-competition}. Furthermore, upweighting the output of each head type reveals a bias towards one of the two answers. Similar to the in-context heads,~\citet{variengien2023look} show that a set of downstream attention heads retrieve the correct answer (an attribute) from the context via copying, preceded by a processing of the request (a question) in middle layers.~\citet{wu2024retrieval} study these type of heads, which they coin \textbf{retrieval heads} in arbitrarily long-contexts, and show they are crucial for solving the Needle-in-a-Haystack tests~\citep{needle_haystack}.~\citet{monea2024glitch} complement the findings of ~\citet{yu-etal-2023-characterizing} and~\citet{meng2022locating} and show that \textbf{FFNs in the last token of the subject have higher contributions on ungrounded (memorized) answers as opposed to grounded answers}, while suggesting that grounding could be a more distributed process lacking a specific localization.~\citet{haviv-etal-2023-understanding} show that the recall of ``memorized'' idioms largely depends on the updates of the FFNs in early layers, providing further evidence of their role as a storage of memorized information. This is further observed in the study of memorized paragraphs, with lower layers exhibiting larger gradient flow~\citep{stoehr2024localizing}. On the other hand,~\citet{sharma2024the} show that substituting the original FFN matrices by lower-rank approximations (\Cref{sec:decoding_model_weights}) leads to improvements in model performance, especially in later layers of the model. They show that, in the factual recall task, the components with smaller singular values encode the correct semantic type of the answer but the wrong answer, thus their removal benefits the accuracy. To conclude, we draw a connection with a decoding strategy (DoLa) proposed to improve the factuality of language models~\citep{chuang2024dola}. DoLa contrastively compares the logit-lens next-token distributions between an early layer and a later layer~\citep{li-etal-2023-contrastive}, promoting tokens that undergo a larger probability change, suggesting that the factual knowledge injection is done in a distributed manner across the network.

\paragraph{Factuality issues and model editing.} Factual information encoded in LMs might be incorrect from the start, or become obsolete over time. Moreover, inconsistencies have been observed when recalling factual knowledge in multilingual and cross-lingual settings~\citep{fierro-sogaard-2022-factual,qi-etal-2023-cross}, or when factual associations are elicited using less common formulations~\citep{Berglund2023TheRC}. This sparked the interest in developing \textit{model editing} approaches able to perform targeted updates on model factual associations with minimal impact on other capabilities. While early approaches proposed edits based on external modules trained for knowledge editing~\citep{de-cao-etal-2021-editing,mitchell2022fast,serac}, recent methods employ causal interventions~(\Cref{sec:causal_intervention}) to localize knowledge neurons~\citep{dai-etal-2022-knowledge} and FFNs in one or more layers~\citep{meng2022locating,memit}, informed by factual recall mechanisms described in the previous paragraph. However, model editing approaches still present several challenges, summarized in~\citep{yao-etal-2023-editing,li2024unveiling}, including the risks of catastrophic forgetting~\citep{Gupta2024ModelEA,gupta2024unified} and downstream performance loss~\citep{Gu2024ModelEC}. Importantly, \citet{hase-etal-2023-localization} show that effective localization does not always result in improved editing results, and that distributed edits across different model sections can result in similar editing accuracy. Steerable-by-design architectures such as the Backpack Transformer~\citep{hewitt-etal-2023-backpack} were recently proposed as possible alternatives to localization-driven methods, exploiting the linearity of component contributions~(\Cref{sec:linear_rep_hypothesis}) as an inductive bias to enhance controllability. We refer readers to~\citet{Wang2023KnowledgeEF} for further insights on LM editing.

\section{LM Interpretability Tools}\label{sec:lm_interpretability_tools}

Several open-source software libraries were introduced to facilitate interpretability studies on Transformer-based LMs. In this section, we briefly summarize the most notable ones and highlight their main points of strength.

\paragraph{Input attribution tools.}  \texttt{Captum}~\citep{kokhlikyan2020captum} is a library in the Pytorch ecosystem providing access to several gradient and perturbation-based input attribution methods for any Pytorch-based model. It notably supports training data attribution methods~(\Cref{sec:training-data-attribution}), and recently added several utilities for simplifying attribution analyses of generative LMs~\citep{miglani-etal-2023-using}. Several \texttt{Captum}-based tools provide convenient APIs for input attribution of Transformer-based models: \texttt{Transformers Interpret}~\citep{transformers-interpret}, \texttt{ferret}~\citep{attanasio-etal-2023-ferret} and \texttt{Ecco}~\citep{alammar-2021-ecco} are mainly centered around language classification tasks, while \texttt{Inseq}~\citep{inseq} is focused specifically on generative LMs and supports advanced approaches for contrastive context attribution~\citep{sarti-etal-2023-quantifying} as well as context mixing evaluation~(\Cref{sec:input_attribution}). \texttt{SHAP}~\citep{lundberg-lee-2017-shap} is a popular toolkit mainly centered on perturbation-based input attribution methods and model-agnostic explanations for various data modalities. The \texttt{Saliency}~\citep{saliency} library provides framework-agnostic implementations for mainly gradient-based input attribution methods. \texttt{LIT}~\citep{tenney-etal-2020-language} is a framework-agnostic tool providing a convenient set of utilities and an intuitive interface for interpretability studies spanning input attribution, concept-based explanations and counterfactual behavior evaluation. It notably includes a visual tool for debugging complex LLM prompts~\citep{tenney2024interactive}.

\paragraph{Component importance analysis tools.} Tools supporting work on circuit discovery and causal interventions play a fundamental role in mechanistic studies, balancing the complexity and model-specific nature of intervention-based methods with a broad support for various pre-trained LM architectures. \texttt{TransformerLens}~\citep{nanda2022transformerlens} is a Pytorch-based toolkit to conduct mechanistic interpretability analyses of generative language models inspired by the closed-source \texttt{Gar\c{c}on} library~\citep{garcon}. The library reimplements popular Transformer LM architectures, preserving compatibility with the popular \texttt{transformers} library~\citep{wolf-etal-2020-transformers} while also providing utilities such as hook points around model activations and attention head decomposition to facilitate custom interventions.~\texttt{NNsight}~\citep{nnsight} provides a Pytorch-compatible interface for interpretability analyses. Its usage is not restricted to Transformer models, but it provides utilities to streamline the usage of \texttt{transformers} checkpoints. Its main peculiarity is the ability to compile an \textit{intervention graph} that can be processed through delayed execution, enabling the extraction of arbitrary internal information from large LMs hosted on remote servers. \texttt{Pyvene}~\citep{wu2024pyvene} is a Pytorch-based library supporting complex intervention schemes, such as trainable~\citep{geiger2023causal} and mid-training interventions~\citep{geiger2022inducing}, alongside various model categories beyond Transformers. Notably, it supports the serialization of intervention schemes to simplify analyses and promotes reusability. Several tools are currently used for the development of SAEs~(\Cref{sec:saes}), providing overlapping sets of features. For example, \texttt{SAELens}~\citep{bloom-channin-2024-saelens} supports advanced visualization of SAE features, while \texttt{dictionary-learning}~\citep{marks-mueller-2023-sae} is an actively developed tool built on top of \texttt{NNsight}, supporting various experimental features to address SAEs' weaknesses. Other examples include~\citet{cooney-2023-sae,wu-etal-2024-sae}, providing standard \texttt{TransformerLens}-compatible SAE implementations, and~\citet{belrose-2024-sae} for Top-K SAE training.

\paragraph{Tools for visualizing model internals.} Several tools such as \texttt{BERTViz}~\citep{vig-2019-multiscale}, \texttt{exBERT}~\citep{hoover-etal-2020-exbert} and \texttt{InterpreT}~\citep{lal-etal-2021-interpret} were developed to visualize attention weights and activations in Transformer-based LMs. \texttt{LM-Debugger}~\citep{geva-etal-2022-lm} is a toolkit to inspect intermediate representation updates through the lens of logit attribution~(\Cref{sec:logit_attribution}), while \texttt{VISIT}~\citep{katz-belinkov-2023-visit}, \texttt{Ecco}~\citep{alammar-2021-ecco} and \texttt{Tuned Lens}\footnote{\url{https://github.com/AlignmentResearch/tuned-lens}}~\citep{belrose2023eliciting} simplify the application of naive and learned vocabulary projections to inspect the evolution of predictions across model layers. \texttt{CircuitsVis}~\citep{circuitvis} provides reusable Python bindings for front-end components that can be used to visualize Transformers internals and predictions, and was adopted by various interpretability tools. \texttt{Penzai}~\citep{johnson-etal-2024-penzai} is a JAX library supporting rich visualizations of pytree data structures, including LM weights and activations. \texttt{LM-TT}~\citep{tufanov2024lm} allows inspecting the information flow in a forward pass, faciliting the examination of the contributions of individual attention heads and feed-forward neurons. \texttt{TDB}~\citep{mossing2024tdb} is a visual interface to interpret neuron activations in LMs supporting automated interpretability techniques and SAEs. \texttt{Neuronpedia}\footnote{\url{https://neuronpedia.org}}~\citep{neuronpedia} provides an open repository for visualizing activation of SAE features trained on LM residual stream states~(\Cref{sec:saes}). Notably, it includes a gamified experience to facilitate the annotation of human-interpretable concepts in SAE feature space. Lastly, \texttt{sae-vis}~\citep{saevis} is a~\texttt{SAELens}-compatible library to produce feature-centric and prompt-centric interactive visualizations of SAE features.

\paragraph{Other notable interpretability-related tools.} The ``Restricted Access Sequence Processing Language'' (\texttt{RASP},~\citealp{rasp}) is a sequence processing language providing a human-readable model for transformer computations. \texttt{Tracr}~\citep{tracr} is a compiler converting RASP programs into decoder-only Transformer weights, automating the creation of small Transformer models implementing specific desired behaviors. \texttt{RASP} and \texttt{Tracr} were adopted for promoting interpretable behaviors via constrained optimization~\citep{friedman2023learning} and validating the effectiveness of circuit discovery techniques~\citep{conmy2023automated}. \texttt{Pyreft}\footnote{\url{https://github.com/stanfordnlp/pyreft}}~\citep{wu2024reft} is a toolkit based on \texttt{Pyvene} for fine-tuning and sharing trainable interventions~(\Cref{sec:circuit_analysis}, Causal abstraction) aimed at optimizing LM performance on selected tasks, in a similar but more targeted and efficient way than parameter-efficient fine-tuning methods (PEFT,~\citealp{han2024parameterefficient}). Going beyond the textual modality, \texttt{ViT Prisma}~\citep{joseph2023vit} is a toolkit to conduct mechanistic interpretability analyses on vision and multimodal models. Finally, \texttt{MAIA}~\citep{shaham2024multimodal} is a multimodal language model augmented with tool use to automate common interpretability workflows such as neuron explanations, example synthesis and counterfactual editing.

\section{Conclusion and Future Directions}

In this paper, we have offered an overview of the existing interpretability methods useful for understanding Transformer-based language models, and have presented the insights they have led to. Although the focus of this work is on practical methods and findings, we acknowledge theoretical studies related to the interpretability of Transformers, such as investigations explaining in-context learning~\citep{akyurek2023what,pmlr-v202-von-oswald23a,xie2022an}, explorations of Transformers through the lens of data compression and representation learning~\citep{yu2023whitebox,voita-etal-2019-bottom}, the study of Transformers' learning dynamics~\citep{tian2024joma,tian2023scan,tarzanagh2024transformers}, or the analyses on their generalization properties on algorithmic tasks~\citep{nogueira2021investigating,anil2022exploring,zhou2024understanding}.

Looking forward, we believe that the ultimate test for insights collected in years of interpretability work remains their applicability in debugging and improving the safety and reliability of future models, providing developers and users with better tools to interact with them and understand the factors influencing their predictions~\citep{Longo2023ExplainableAI}. To ensure such requirements are met, future developments in interpretability research will be faced with the challenging task of moving from \textit{functionally-grounded evaluations} (i.e. no human evaluation, only toy settings) to actionable insights and benefits for real-world tasks~\citep{doshivelez2017rigorous}. From an analytical standpoint, this involves moving from methods and analyses operating in model component space to human-interpretable space, i.e from model components to features and natural language explanations, as suggested by~\citet{singh-etal-2024-rethinking}, while still faithfully reflecting model behaviors~\citep{siegel2024probabilities}. Directions we deem promising in this area involve the usage of LMs as ~\textit{verbalizers}~\citep{feldhus-etal-2023-saliency,openai_neuron_nle,wang2024llmcheckup,chen2024selfie} for scaling input and component attribution analyses, especially when paired with verification mechanisms to ensure counterfactual consistency~\citep{Avitan2024WhatCC}, and circuit discovery methods leveraging interpretable features to enable interventions motivated by human-understandable concepts~\citep{marks-etal-2024-feature}. More accessible insights might also unlock gains in model performance and efficiency, translating interpretability-driven insights into downstream task improvements~\citep{wu2024reft}. Importantly, interdisciplinary research grounded in the technical developments we summarize in this survey will play a key role in broadening the scope of interpretability analyses to account for the perceptual and interactive dimensions of model explanations from a human perspective~\citep{questioning-ai,who-what-when-xai,xai-overreliance}. Ultimately, we believe that ensuring open and convenient access to the internals of advanced LMs will remain a fundamental prerequisite for future progress in this area~\citep{ndif,casper-etal-2024-blackbox,hudson2024trillion}.

\section*{Acknowledgements}

Javier Ferrando is supported by the Spanish Ministerio de Ciencia e Innovación through the project PID2019-107579RB-I00 / AEI / 10.13039/501100011033. Gabriele Sarti and Arianna Bisazza acknowledge the support of the Dutch Research Council (NWO) as part of the project InDeep (NWA.1292.19.399). The authors express their gratitude to Paul Riechers, Dmitrii Troitskii and Alex Loftus for their useful comments.

\footnotesize
%\setcitestyle{numbers}
\bibliography{anthology,custom}
\bibliographystyle{iclr2023_conference}
\newpage

\appendix
\normalsize
\section{Mathematical Notation}\label{apx:notation}
%\begin{comment}
\begin{table*}[!h]
\resizebox{\textwidth}{!}{%
\centering
\begin{tabular}{cc}
\toprule
Notation & Definition\\
\midrule
$n$         & Sequence length \\
$\mathcal{V}$         & Vocabulary \\
$\mathbf{t} = \langle t_1, t_2 \ldots, t_n \rangle$         & Input sequence of tokens \\
$\mathbf{x} = \langle \vx_1, \vx_2 \ldots, \vx_n \rangle$         & Input sequence of token embeddings \\
$d$         &  Model dimension \\
$d_h$         &  Attention head dimension \\
$d_{\text{FFN}}$         &  FFN dimension \\
$H$                    & Number of heads             \\
$L$                    & Number of layers             \\
$\vx^{l}_{i} \in \mathbb{R}^{d}$         &  Residual stream state at position $i$, layer $l$ \\
$\vx^{\text{mid},l}_{i} \in \mathbb{R}^{d}$         &  Residual stream state at position $i$, layer $l$, after the attention block \\
%$\mathcal{C}_{i}$         &  Set of components at position $i$ \\
$f^c(\mathbf{x}) \in \mathbb{R}^{d}$         &  Component $c$ output representation at the last position\\
$f^l(\mathbf{x}) = \vx^{l}_{n} \in \mathbb{R}^{d}$         &  Residual stream state at the last position, layer $l$\\
$\mA^{l,h} \in \mathbb{R}^{n \times n}$             & Attention matrix at layer $l$, head $h$                           \\
$\mW_Q^{l,h}, \mW_K^{l,h}, \mW_V^{l,h} \in \mathbb{R}^{d \times d_h}$         & Queries, keys and values weight matrices at layer $l$, head $h$                      \\
$\mW_O^{l,h} \in \mathbb{R}^{d_h \times d}$         & Output weight matrix at layer $l$, head $h$                      \\
$\mW_{\text{in}}^{l} \in \mathbb{R}^{d \times d_{\text{FFN}}}$, $\mW_{\text{out}}^{l} \in \mathbb{R}^{d_{\text{FFN}} \times d}$         & FFN input and  output weight matrices at layer $l$                   \\
$\mW_{E} \in \mathbb{R}^{d \times |\mathcal{V}|}$ and $\mW_{U} \in \mathbb{R}^{|\mathcal{V}| \times d}$        & Embedding and unembedding matrices                   \\
\bottomrule
\end{tabular}
}
\caption{Notation and definitions of the main variables used in this work.}
\end{table*}

%\end{comment}
\section{Linearization of the LayerNorm}\label{apx:layernorm_linear}
The LayerNorm operates over an input $\vz$ as: $\text{LN}(\vz)=\frac{\vz-\mu(\vz)}{\sigma(\vz)} \odot \mathbf{\gamma}+ \mathbf{\beta}$, where $\mu$ and $\sigma$ compute the mean and standard deviation of $\vz$, and $\gamma$ and $\beta$ refer to the element-wise transformation and bias respectively.
Holding $\sigma(\vz)$ as a constant, the LayerNorm can be decomposed into $\vz\mathbf{L}+ \beta$, where $\mathbf{L}$ is a linear transformation:
\begin{equation}\label{eq:linear_layernorm}
\resizebox{0.48\textwidth}{!}{$\displaystyle{
\mathbf{L}:=\frac{1}{\sigma(\vz)}\left[ 
\begin{array}{cccc}
\gamma _{1} & 0 & \cdots  & 0 \\ 
0 & \gamma _{2} & \cdots  & 0 \\ 
\cdots  & \cdots  & \cdots  & \cdots  \\ 
0 & 0 & \cdots  & \gamma _{d}%
\end{array}%
\right] \left[ 
\begin{array}{cccc}
\frac{d-1}{d} & -\frac{1}{d} & \cdots  & -\frac{1}{d} \\ 
-\frac{1}{d} & \frac{d-1}{d} & \cdots  & -\frac{1}{d} \\ 
\cdots  & \cdots  & \cdots  & \cdots  \\ 
-\frac{1}{d} & -\frac{1}{d} & \cdots  & \frac{d-1}{d}%
\end{array}%
\right]}$}.
\end{equation}%

%The linear map on the right subtracts the mean to the input vector, $\vz' = \vz-\mu(\vz)$. The left matrix performs the hadamard product with the layer normalization weights ($\vz' \odot \gamma$).
\section{Folding the LayerNorm}
\label{appx:folding_ln}

Any Transformer block reads from the residual stream by normalizing before applying a linear layer (with weights $\mW$ and bias $\vb$) to the resulting vector:
\begin{equation}
    \text{LN}(\vz)\mW + \vb
\end{equation}

Following the decomposition in \Cref{eq:linear_layernorm} we can fold the weights of the LayerNorm into those of the subsequent linear layer as follows:
\begin{align*}\label{eq:folded_ln}
\text{LN}(\vz)\mW + \vb &= (\vz \mL + \beta)\mW + \vb\\
&= \vz\mL\mW + \beta \mW + \vb\\
&= \vz\mW^* + \vb^*\numberthis,
\end{align*}
where the new weights and bias are $\mW^* = \mL\mW$ and $\vb^* = \beta \mW + \vb$ respectively.

\section{Implementation details of SAEs}\label{apx:imple_details_saes}
\begin{itemize}
    \item During training, a feature receives a zero gradient signal if it does not activate. When this occurs frequently, it can lead to a dead feature.~\citet{bricken2023monosemanticity} propose \textbf{resampling} these features by reinitializing their encoder and decoder weights periodically during training. An alternative approach to resampling is \textbf{ghost gradients}~\citep{anthropic_january_update_ghost_grads}, which adds an auxiliary loss term that supplies a gradient signal to promote the reactivation of dead features. However, recent results have found this approach suboptimal~\citep{update_1_gdm_mech_interp_ghost_grads,anthropic_april_update}. 
    %However, resampling has been found to cause an increase in L0 norms and loss recovered~\citep{rajamanoharan2024improving,guess_saes}.
    \item Setting the $\beta_1$ parameter of Adam to 0 has been found to reduce the number of ``dead'' features in larger autoencoders~\citep{anthropic_february_update,rajamanoharan2024improving}. Yet, \citet{anthropic_april_update} rely on $\beta_1=0.9$.
    \item Although intially the \textbf{norm of the decoder's rows}\footnote{Note that we consider the decoder weight matrix ${\mW_{\text{dec}} \in \mathbb{R}^{m \times d}}$.} was recommended to be equal to one~\citep{bricken2023monosemanticity}, recent released SAEs also consider an unconstrained norm setting~\citep{anthropic_april_update}.
    
\end{itemize}

\end{document}